\documentclass[final,3p,times,twocolumn]{elsarticle}

\usepackage{amssymb}
\usepackage{hyperref}
\usepackage{graphicx}
\usepackage{subcaption}
\usepackage{lscape}
\usepackage{amsmath}
\usepackage{multirow}
\usepackage[figuresright]{rotating}
\usepackage{lineno}
\usepackage{algorithm} 
\usepackage{algpseudocode} 
\usepackage{comment}
\usepackage{lineno}
\usepackage{nomencl}
\makenomenclature
\usepackage{calc}

\usepackage{natbib}
\setcitestyle{authoryear,open={(},close={)},citesep={;}} 

\usepackage{calc}
\begin{document}

\begin{frontmatter}



\title{Generative Adversarial Networks for Image Augmentation in Agriculture: \\A Systematic Review}

\author[label1]{Ebenezer Olaniyi}
\author[label2]{Dong Chen}
\author[label1]{Yuzhen Lu}
\author[label3]{Yanbo Huang}

\address{* E.O, D.C. and Y.L. contributed equally to the review}
\address{Yuzhen Lu (yl747@msstate.edu) is the corresponding author}

\address[label1]{Department of Agricultural and Biological Engineering, Mississippi State University, Mississippi State 39762, MS, USA}
\address[label2]{Department of Electrical and Computer Engineering, Michigan State University, East Lansing, MI 48824, USA}
\address[label3]{United States Department of Agriculture, Agricultural Research Service, Genetics and Sustainable Agriculture Research Unit, Mississippi State, MS 39762, USA}

\begin{abstract}
In agricultural image analysis, optimal model performance is keenly pursued for better fulfilling visual recognition tasks (e.g., image classification, segmentation, object detection and localization), in the presence of challenges with biological variability and unstructured environments. Large-scale, balanced and ground-truthed image datasets, however, are often difficult to obtain to fuel  the development of advanced, high-performance models. As artificial intelligence through deep learning is impacting analysis and modeling of agricultural images, data augmentation plays a crucial role in boosting model performance while reducing manual efforts for data preparation, by algorithmically expanding training datasets. Beyond traditional data augmentation techniques, generative adversarial network (GAN) invented in 2014 in the computer vision community, provides a suite of novel approaches that can learn good data representations and generate highly realistic samples. Since 2017, there has been a growth of research into GANs for image augmentation or synthesis in agriculture for improved model performance. This paper presents an overview of the evolution of GAN architectures followed by a systematic review of their application to agriculture (\url{https://github.com/Derekabc/GANs-Agriculture}), involving various vision tasks for plant health, weeds, fruits, aquaculture, animal farming, plant phenotyping as well as postharvest inspection of fruits. Challenges and opportunities of GANs are discussed for future research. 

\end{abstract}

\begin{keyword}
GAN, Image Augmentation, Agriculture, Food, Deep Learning, Domain Adaptation
\end{keyword}

\end{frontmatter}

\nomenclature{AC-GAN}{Auxiliary classifier generative adversarial network}
\nomenclature{AdaIN}{Adaptive instance normalization }
\nomenclature{ADA}{Adversarial domain adaptation}
\nomenclature{AI}{Artificial intelligence}
\nomenclature{AP}{Average precision}
\nomenclature{AR-GAN}{Attentive recurrent generative adversarial network}
\nomenclature{BN}{Batch normalization}
\nomenclature{CGAN}{Conditional generative adversarial network}
\nomenclature{CNN}{Convolutional neural network}
\nomenclature{CoGAN}{Coupled generative adversarial network}
\nomenclature{DCGAN}{Deep convolutional generative adversarial network}
\nomenclature{DL}{Deep learning}
\nomenclature{ESR-GAN}{Enhanced super-resolution generative adversarial network}
\nomenclature{FCN}{Fully connected network}
\nomenclature{FID}{Fréchet inception distance}
\nomenclature{GAN}{Generative adversarial network}
\nomenclature{InfoGAN}{Information maximizing generative adversarial network}
\nomenclature{IoU}{Intersection over union}
\nomenclature{mAP}{Mean average precision}
\nomenclature{mIoU}{Mean intersection over union}
\nomenclature{NIR}{Near-infrared}
\nomenclature{PCA}{Principal component analysis}
\nomenclature{PNAS}{Progressive neural architecture search}
\nomenclature{ProGAN}{Progressively growing generative adversarial network}
\nomenclature{ReLU}{Rectified linear unit }
\nomenclature{RGB}{Red-Green-Blue}
\nomenclature{SA-GAN}{Self-attention generative adversarial network}
\nomenclature{SR-GAN}{Super-resolution generative adversarial network}
\nomenclature{StyleGAN}{Style generative adversarial network}
\nomenclature{SVM}{Support vector machine}
\nomenclature{TransGAN}{Transformer generative adversarial network}
\nomenclature{UAV}{Unmanned aerial vehicle}
\nomenclature{WGAN}{Wasserstein generative adversarial network}
\nomenclature{WANG-GP}{Wasserstein generative adversarial network with gradient penalty}
\printnomenclature[0.7in]

\section{Introduction}
\label{sec:intro}
Machine vision or imaging technologies are ubiquitous in agricultural applications. A diversity of imaging techniques, e.g., panchromatic/monochromatic, color or RGB (red-green-blue), near-infrared (NIR) multispectral/hyperspectral imaging, are available to replace human vision to acquire information from agricultural processes and food products \citep{chen2002machine, davies2009application, mavridou2019machine, lu2020hyperspectral}, offering improved efficacy, efficiency, and objectivity in vision tasks relevant to agriculture (e.g., crop stress diagnosis, weed control, phenotyping, robotic fruit harvesting, and grading and sorting of agricultural produce). Integrated with machine vision technologies, machine learning offers a suite of methods/algorithms \citep{liakos2018machine, rahman2019machine} for analyzing and learning patterns from data and has been widely utilized in data-driven visual recognition (e.g., object classification, detection/localization, and segmentation) and quantification tasks (e.g., plant stress quantification) in agriculture-food systems. Traditional machine learning methodologies for image analysis require the extraction of hand-engineered features (e.g., color, shape, texture) from images to build pattern recognition and classification models \citep{o2019deep}, which is a cumbersome, trial-and-error process without guaranteeing optimal model performance.  

In the past ten years, with the advancements in computing hardware including GPUs, artificial intelligence (AI) with deep learning (DL) \citep{bengio2021deep, lecun2015deep, krizhevsky2012imagenet} has emerged as a new paradigm and workhorse for pattern classification within the field of artificial intelligence or machine learning. Popularized by supporting frameworks \citep{DBLP:journals/corr/abs-1912-01703, 199317}, DL is advantageous as it enables seamless integration of feature extraction and pattern classification processes while learning high-quality image representations and achieving state-of-the-art performance in visual recognition tasks \citep{krizhevsky2012imagenet, he2016deep, carion2020end, DBLP:journals/corr/abs-2106-04560}, which otherwise could be challenging using traditional machine learning methods. The impacts of DL, especially convolutional neural networks (CNNs), on tackling machine vision-guided agricultural tasks, have been well reviewed in \citet{kamilaris2018deep}, \citet{koirala2019deep} and \cite{zhang2020applications}. Despite impressive results in analyzing and modeling agricultural images, it has been clearly demonstrated that supplies of large-scale datasets are necessary for ensuring the performance of DL \citep{sun2017revisiting} or advanced machine learning models \citep{paullada2021data} while avoiding overfitting. To achieve human-level performance, as a rough rule of thumb, thousands of images per category for achieving are expected to train DL models from scratch \citep{lecun2015deep, Goodfellow-et-al-2016}. Given the positive impact of data size, a number of immensely large image datasets with millions of labeled images have been created in the computer vision community, e.g., ImageNet \citep{deng2009imagenet}, COCO \citep{lin2014microsoft}, LVIS \citep{gupta2019lvis}, Open Images \citep{OpenImages}, just to name a few.  

Analysis and modeling of agricultural images face a myriad of challenges resulting from significant biological variability (no two leaves are the same) and unstructured environments (e.g., object occlusion, variable lighting conditions, cluttered scenes) \citep{bechar2016agricultural, barth2020optimising, vougioukas2019agricultural}. To address these challenges for building high-performance robust models accentuates the need for large-scale datasets that capture sufficient variations under a wide range of imaging conditions. Collecting and annotating such big datasets with ground-truths, however, are time-consuming, resource-intensive, and costly. This is especially true for specific applications (e.g., plant disease detection, weed recognition, fruit defect detection) that place constraints on biological materials and imaging conditions and for precise annotations at pixel level. Currently, there are very few image datasets that consist of hundreds if not thousands of images per category, on par with the aforementioned datasets in computer vision, and are publicly available \citep{lu2020survey}. The scarcity of annotated, large-scale image datasets relevant to agricultural processes is posing a crucial bottleneck to harnessing the power of advanced AI including DL algorithms in the agriculture and food domain. One common approach to mitigating the insufficiency of physically collected data is data/image augmentation that algorithmically expands the scale and variations of datasets \citep{lu2020survey}. A suite of image augmentation techniques \citep{simard2003best, wong2016understanding, shorten2019survey, khalifa2021comprehensive} has been proposed for improving the performance of DL models. 

Collectively, there are two categories of image augmentation techniques, including basic image augmentation and the augmentation based on DL algorithms. The former involves various image transformations using geometric or color processing approaches \citep{shorten2019survey}. The second is achieved using DL techniques \citep{khalifa2021comprehensive}, among which generative adversarial networks (GANs) offer a novel method for image augmentation through generative model learning of underlying distributions of training data. In supervised machine learning/DL pipelines, the original dataset is initially partitioned into training, validation, and testing sets (or training and testing sets). It should be noted that image augmentation is normally conducted for training (and or validation) images alone for model training and optimization \citep{lecun2015deep}, with test data un-augmented to avoid data leakage, although image augmentation is found to be useful at test time for assessing model performance \citep{shanmugam2022sustainable}.  

Originally proposed by \cite{goodfellow2014generative}, GAN is a new framework of generative modeling \citep{Tomczak2022}, which aims to synthesize new data with the same characteristics of training instances (usually images), visually resembling the data in the training set. Since its advent, GAN has enjoyed great popularity for generating realistic data, and various GAN-based methods have been proposed for image syntheses in the past few years \citep{creswell2018generative, huang2018introduction, yi2019generative, cui2021gan}, with applications spreading rapidly from computer vision and machine learning communities to domain-specific areas such as biomedical diagnosis \citep{DBLP:journals/corr/abs-1809-07294, bissoto2021gan}, bioinformatics \citep{lan2020generative} as well as agriculture. While traditional image augmentation techniques are commonly used in model development \citep{simard2003best, cirecsan2010deep}, the diversity or variations garnered through simple geometric/color transformations can be small with little additional information. Instead, GANs provide novel and promising ways through representation learning to potentially induce more variations and enrich datasets, benefiting downstream modeling tasks \citep{perez2017effectiveness, bowles2018gan,frid2018gan}. While a few articles describe theoretical aspects and generic applications of GANs \citep{lecun2015deep, DBLP:journals/cacm/GoodfellowPMXWO20,huang2018introduction,gui2021review,creswell2018generative, yi2019generative, liu2021generative}, none of them, to the best of our knowledge, are specifically devoted to the application of GANs for image augmentation in the agricultural and food domain.  

This paper is therefore to provide the first systematic review of GANs for image augmentation/synthesis in agriculture to facilitate DL-assisted computer/machine vision tasks. We survey recent developments of GAN architectures and their application in precision agriculture, phenotyping, and postharvest handling, and discuss technical challenges facing GANs and future research needs. This review would be beneficial for the research communities to exploit GAN techniques to address challenges from the shortage of labeled, large image datasets in agriculture and food systems. 

The remainder of the paper is organized as follows. Section~\ref{sec:Methods} describes the method and exclusion criteria for the literature review. Section~\ref{sec:basic_image_aug} describes basic image augmentation methods for machine learning systems. Described next is a comprehensive overview of representative GAN architectures in Section ~\ref{sec:gans}. A systematic review of agricultural applications of GANs is presented in Section~\ref{sec:app_gans}, followed by a discussion of challenges and further work of GANs in  Section~\ref{sec:dis} and a conclusive summary in Section~\ref{sec:conclu}.

\section{Methods}
\label{sec:Methods}
\begin{figure*}[!ht]
  \centering
  \includegraphics[width=0.7\textwidth]{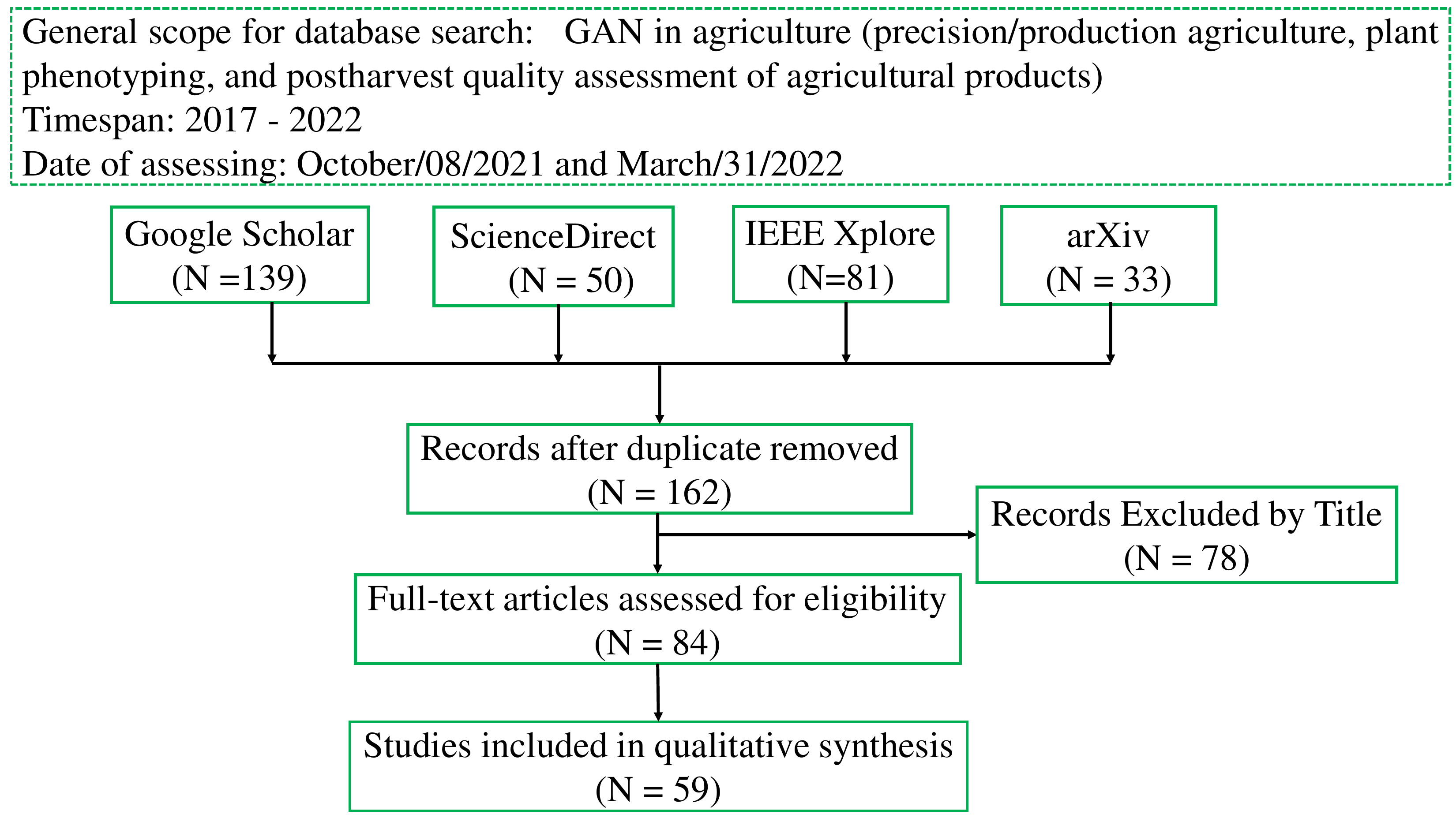}
  \caption{The PRISMA guideline flowchart used in this review. The Figure first row illustrates the initial selected articles based on the keywords that enhanced the initial filtering before other exclusion criteria are applied.}
  \label{fig1}
\end{figure*}

This review adopts the PRISMA (preferred reporting items for systematic reviews and meta-analysis) guideline \cite{moher2009preferred}, an evidence-based framework to perform a systematic review. Fig.~\ref{fig1} shows the PRISMA guideline flowchart. All the authors agreed upon the design and method that follows the PRISMA guideline for this review. Likewise, all the authors agreed on the focused questions, search databases, inclusion and exclusion criteria for the articles, and the PRISMA flowchart.  

The focused questions considered in this review include: can new training images be generated to expand the training data? The significance of image augmentation in DL for agricultural applications? How can GANs help to expand training data for improving DL model performance?  Regarding information sources, the literature search was conducted following the PRISMA guideline \citep{moher2009preferred}. We initially searched several scientific databases (e.g., Web of Science, ScienceDirect, Google Scholar, Springer) to collate the literature. We discovered that some databases cover limited articles since GAN in agriculture is an emerging application area. Articles covered in one database with limited articles were also covered by another database with broader coverage. Therefore, we discarded the databases with limited and repeated articles. Finally, three mainstream scientific journal databases including Google Scholar, ScienceDirect and IEEE Xplore, along with the open-access arXiv database, were considered in this review. Relevant articles were searched and chosen using the search strategy and inclusion criteria described next. 

To collect relevant literature, we searched the databases based on selected keywords, which were entered into the search engine of each database. We first did searching with individual keywords such as ``synthetic images of plant diseases", ``weed detection data augmentation", ``synthetic images of weeds", ``plant disease data augmentation", ``data augmentation in agriculture", ``generative adversarial network in agriculture", etc. To collect more publications, we also grouped similar keywords with AND and the keywords in diverse groups with OR. This resulted in search strings such as ``synthetic images of weeds" AND ``artificial images in weed control", ``generative adversarial network in pest recognition" OR ``synthetic images in plant seedlings" , ``artificial images in postharvest" OR ``aquaculture synthetic images", ``image augmentation in plant seedling" OR ``artificial images in weed control", ``generative adversarial network in fruit detection" AND ``synthetic images in fruit detection", to name a few.

\begin{figure*}[!ht]
  \centering
  \includegraphics[width=0.4\textwidth]{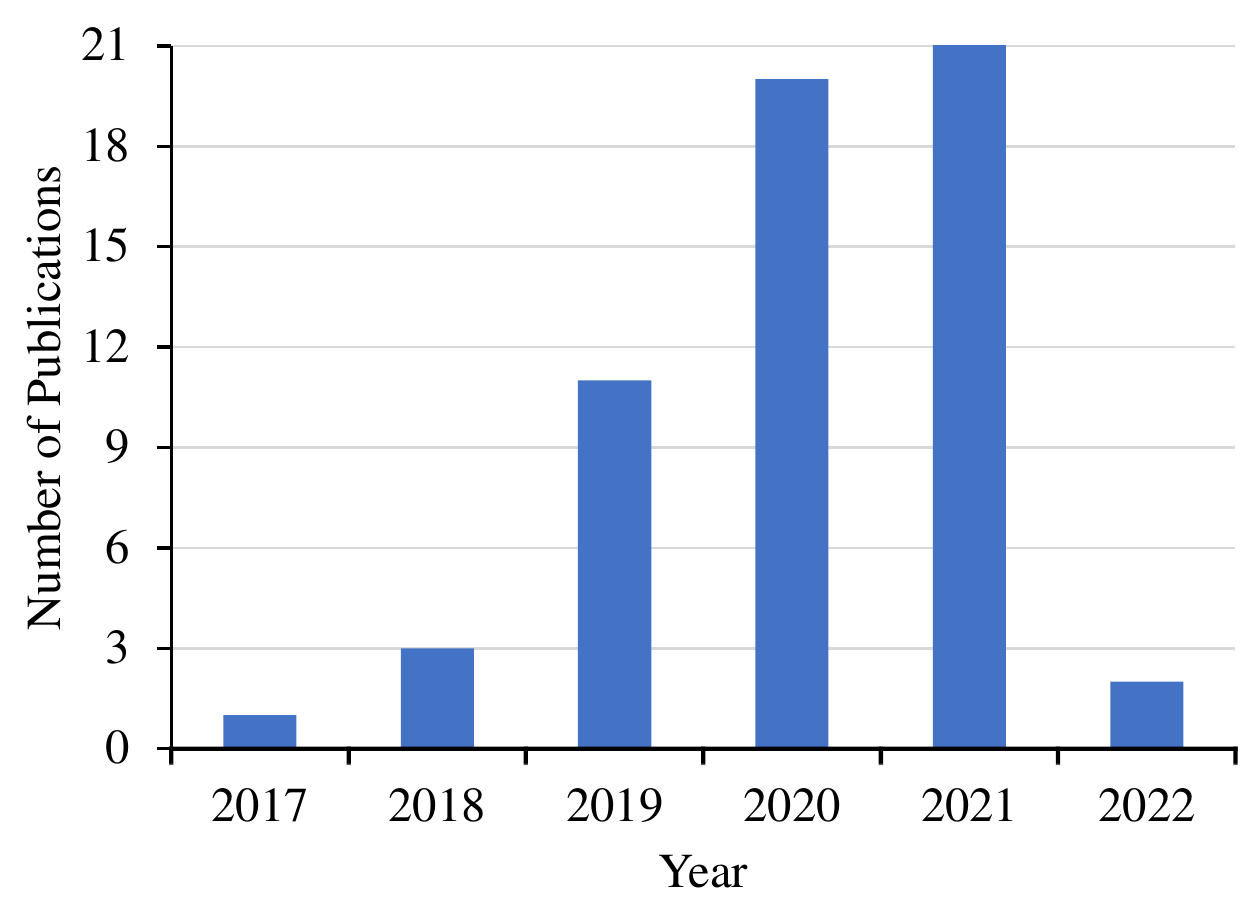}
  \caption{Numbers of publications over the years.}
  \label{fig2}
\end{figure*}

We restricted our search to only articles written in English and published within five past years (2017-2022) due to the recent emergence of GANs in the agricultural field. Finally, we collected 303 articles from the database search, comprising 139 articles from Google Scholar, 50 articles from ScienceDirect, 81 articles from IEEE Xplore and 33 articles from arXiv. Articles screening was then carried out based on duplicate, title, abstract, and full text of the papers using the following exclusion criteria: (1) review articles, (2) articles irrelevant to data augmentation in agriculture, (3) publications that are duplicate or already enlisted from another search database, and (4) publications that do not include the full text. 

After the eligibility and exclusion criteria were thoroughly implemented, we were left with a total number of 59 articles at the time of writing (Fig.~\ref{fig1} and Fig.~\ref{fig2}). The articles reviewed in this study are summarized on our Github repository\footnote{\url{https://github.com/Derekabc/GANs-Agriculture}}, which will be actively maintained.

\section{Basic Image Augmentation}
\label{sec:basic_image_aug}
Basic image augmentation is loosely referred to as transformations that manipulate the geometry and color characteristics of images \citep{taylor2018improving,khalifa2021comprehensive}, corresponding to geometric and photometric transformations, respectively. The geometric transformations (also known as image warping) involve altering the geometry and sometimes associated spatial information (e.g., masks, bounding boxes) of an image. The object shape in the image may be preserved or not, depending on the linearity of transformation methods. Affine transformations that preserve spatial collinearity and are traditionally used in image registration and mosaicking \citep{shapiro2001computer}, are most often used for image augmentation when training deep learning models, including translation, rotation, magnification, and composition of any combination and sequence. There are also non-linear warping methods \citep{oliveira2017augmenting}, e.g., elastic deformations in \cite{simard2003best}. The photometric methods involve the transformations across color channels of an RGB image \cite{taylor2018improving}, or spectral channels for multi-/hyper-spectral images. These methods do not change object geometry and spatial characteristics. Color transformations are commonly done through color jittering that is to randomly adjust color saturation, brightness, and contrast in image color spaces \citep{afifi2019color,cubuk2019autoaugment}, color channel swapping, principal component analysis (PCA) color augmentation (also known as Fancy PCA) \citep{krizhevsky2012imagenet, taylor2018improving}, among others. Common image processing methods, such as filtering, histogram equalization, edge enhancement, noise addition, etc., which have the effect of inducing variations to image characteristics (e.g., shape, color and or texture), can also be used for image augmentation.

\begin{figure*}[!ht]
  \centering
  \includegraphics[width=0.7\textwidth]{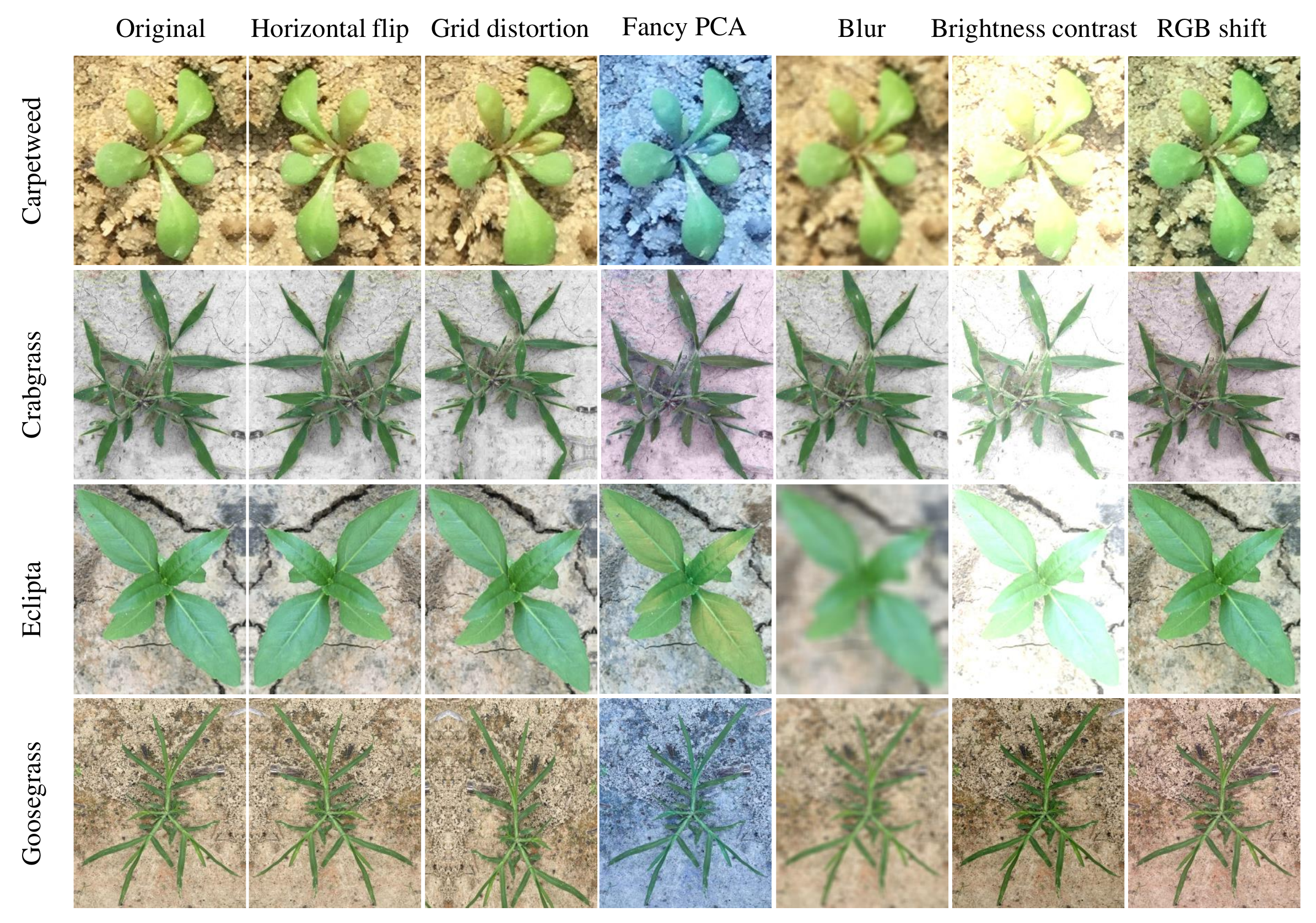}
  \caption{Illustration of basic image augmentation on weed images.}
  \label{fig3}
\end{figure*}

Fig.~\ref{fig3} illustrates common basic image augmentation methods for four weed images, randomly chosen in the CottonWeed dataset \citep{chen2021performance}. The augmentation was achieved using Albumentation \citep{buslaev2020albumentations}, an open-source library for image augmentation. Although DL frameworks, such as TensorFlow and PyTorch, have the functionality of basic image augmentation, a number of standalone, software packages dedicated to image augmentation tasks using a rich variety of spatial/pixel-level transformation operations, have been developed and publicly available to enhance image augmentation. Table~\ref{tab1} summarizes common open-source packages/tools for image augmentation. 

\begin{table*}[!ht]
\renewcommand{\arraystretch}{1.2}
\caption{Example of open-source software packages for image augmentation.}
\label{tab1}
\centering
\resizebox{0.7\textwidth}{!}{
\begin{tabular}{l|l|ll}
\hline
Packages/Tools &   URL     & \multicolumn{2}{l}{Reference(s)}                                 \\ \hline
Albumentations &  \url{https://github.com/albumentations-team}     & \multicolumn{2}{l}{ \cite{buslaev2020albumentations}} \\ \hline
Augmentor      &  \url{https://github.com/mdbloice/Augmentor}      & \multicolumn{2}{l}{ \cite{bloice2019biomedical}}  \\ \hline
CutBlue        & \url{https://github.com/clovaai/cutblur} & \multicolumn{2}{l}{\cite{yoo2020rethinking}}  \\ \hline
Gryds          & \url{https://github.com/tueimage/gryds} & \multicolumn{2}{l}{ \cite{eppenhof2019progressively}}      \\ \hline
imgaug         &      \url{https://github.com/aleju/imgaug}    & \multicolumn{2}{l}{N/A   }     \\ \hline
Pymia          & \url{https://github.com/rundherum/pymia} & \multicolumn{2}{l}{ \cite{jungo2021pymia} }  \\ \hline
Rising         &     \url{https://github.com/PhoenixDL/rising}   & \multicolumn{2}{l}{N/A } \\ \hline
TorchIO        & \url{https://github.com/fepegar/torchio} & \multicolumn{2}{l}{ \cite{perez2021torchio}}     \\ \hline
\end{tabular}
}
\end{table*}

While basic augmentation methods described above can produce significant amounts of data and help train DL models for improved performance, these methods still have their drawbacks. The basic image transformations (e.g., translation, cropping, rotation) tend to generate highly correlated samples and may not account for variations resulting from different imaging protocols or sequences, which are thus not guaranteed to be effective for model generalization or even can lead to overfitting \citep{shorten2019survey}. Generally, these augmentation techniques are applied on one image at a time and thus inadequate to learn the variations or invariant features from the rest of training data. Sometimes, heavy augmentations may eliminate or damage meaningful semantic contents, which is undesirable for images acquired under strict standards \citep{konidaris2019generative, yi2019generative}. Little new information would be gained from otherwise small modifications to the images (e.g., translation of an image by a few pixels, image cropping). GANs, which are described in the following sections, offer a potentially valuable addition to the arsenal of basic image augmentation techniques available, which have the ability to unlock the additional information and enable more variability for further improving the performance of DL models.

\section{Principle and Architectures of Generative Adversarial Networks (GANs)}
\label{sec:gans}
\begin{figure*}[!ht]
  \centering
  \includegraphics[width=0.6\textwidth]{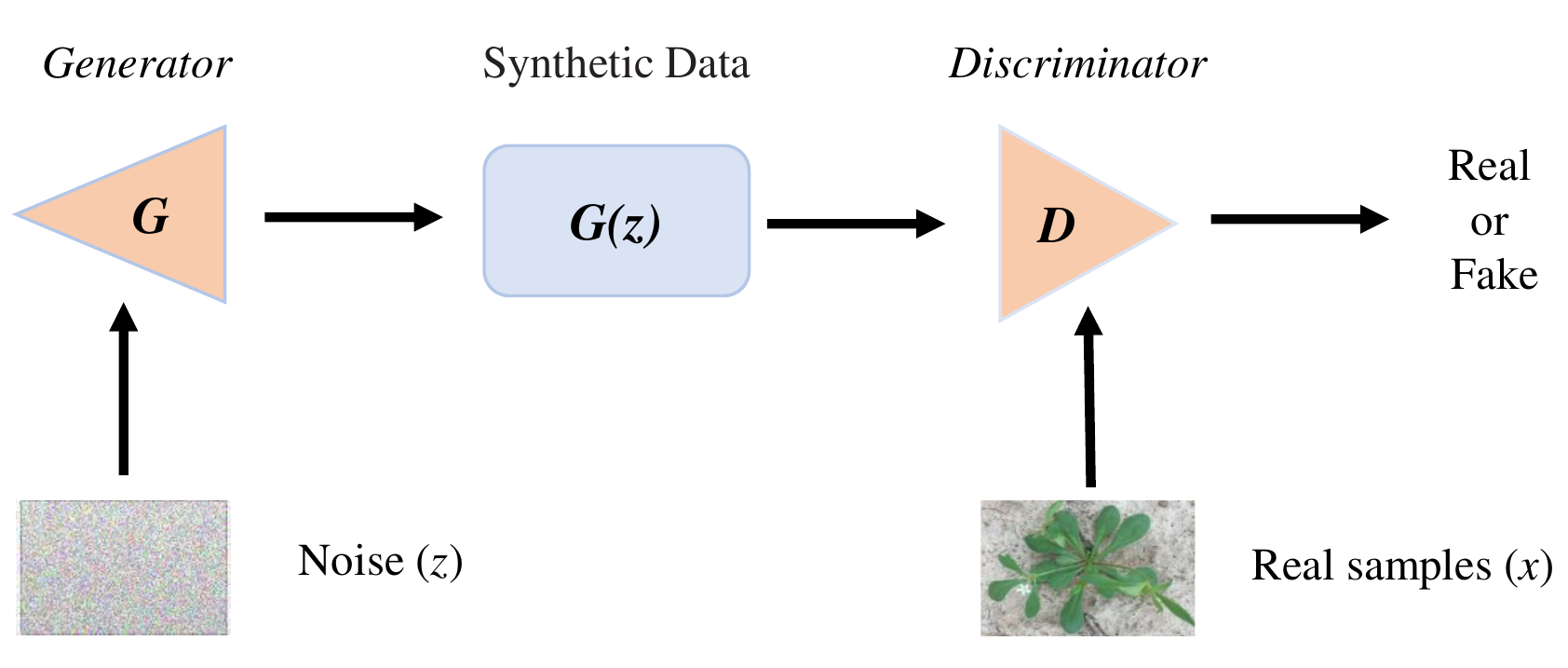}
  \caption{The framework of a vanilla generative adversarial network (GAN) that consists of two main models, i.e., the \textit{generator} ($G$) and the \textit{discriminator} ($D$).}
  \label{fig4}
\end{figure*}

As an emerging class of unsupervised modeling techniques, GANs \citep{goodfellow2014distinguishability, DBLP:journals/cacm/GoodfellowPMXWO20} has brought breakthroughs to data generation for deep learning. The vanilla GAN \citep{goodfellow2014distinguishability} consists of two machine learning models (generally neural networks) called \textit{generator} ($G$) and \textit{discriminator} ($D$), which are trained in an adversarial process. As illustrated in Fig.~\ref{fig4}, $G$ is fed with random noise ($z$) and outputs synthetic data, while $D$ fed with real samples ($x$) is to discriminate between the real and fake samples, $G(z)$, generated from the $G$. Simply put, $G$ is trained to trick the $D$, while the $D$ is optimized not to be fooled by the $G$.  

The original way to train a GAN model is to find a discriminator with the maximized classification and a \textit{generator} that maximally confuses the \textit{discriminator}. The training process involves optimizing the following value function \citep{goodfellow2014distinguishability}: 

\begin{equation} \label{eqn1}
    V(D,G) = E_{x \sim p_x}[\log D(x)] + E_{z \sim p_z}[\log (1-D(G(z)))],
\end{equation}
where $p_x$ and $p_z$ denote the real data and generated data distributions, respectively. The $D$ is trained to maximize the probability of assigning the correct labels to fake samples from the $G$ as well as training samples, and simultaneously $G$ is trained to minimize the loss $log(1 - D(G(z)))$, which leads to the two-player minimax game \citep{goodfellow2014distinguishability}. During training, $G$ and $D$ models are updated iteratively, with the parameters of one model updated and the parameters of the other fixed. When the two models $G$ and $D$ are fully trained, the game reaches a Nash equilibrium \citep{mirza2014conditional, cao2018recent} where $D$ is unable to distinguish the samples from the real data distribution or generated distribution from $G$, thus always predicting 0.5 for all samples.  

Training the standard GAN \citep{goodfellow2014distinguishability} is well known for being unstable \citep{arjovsky2017towards}. Different objective or loss functions have been proposed for stabilized training of GANs, such as \textit{f}-divergence \citep{nowozin2016f}, least-square \citep{mao2017adaptive}, Wasserstein distance (also called Earth-Mover distance) \citep{arjovsky2017towards} and hinge loss \citep{miyato2018spectral}. Among these is the Wasserstein metric arguably the most popular for measuring the distance between real and generated samples, leading to a new GAN framework, called Wasserstein GAN (WGAN) \citep{arjovsky2017towards}. Compared to the original GAN, WGAN has better theoretical properties and improved stability of learning while providing meaningful learning curves to facilitate hyperparameter selection and debugging. Further improvements were made by \cite{gulrajani2017improved} who proposed WGAN with gradient penalty (WGAN-GP) by penalizing the norm of \textit{discriminator} gradients with respect to data input, instead of clipping the weight parameters of the \textit{discriminator} in the original WGAN.   

The networks $D$ and $G$ are often parameterized by differentiable functions which can be fully connected, convolutional or recurrent networks. The original GAN \citep{goodfellow2014distinguishability} used fully connected networks as its building block, which are only suitable for handling low-resolution images datasets, e.g., MNIST and CIFAR-10, since fully connected networks have limited image representation capacity. To synthesize high-resolution, complex images, convolutional neural networks (CNNs) that have more powerful representation ability have become core components in recent GAN models, providing better performance in image generation. Extended from the vanilla GAN model \citep{goodfellow2014distinguishability}, a myriad of GAN architecture variants has been proposed in literature for different applications \citep{wang2020deep,huang2018introduction, creswell2018generative}. Presented below are several important GAN architectures that have been used to generate synthetic images in agricultural applications as well as other disciplines (e.g., computer vision, medical imaging, biology) to improve model performance. 

\begin{figure*}[!ht]
  \centering
  \includegraphics[width=0.6\textwidth]{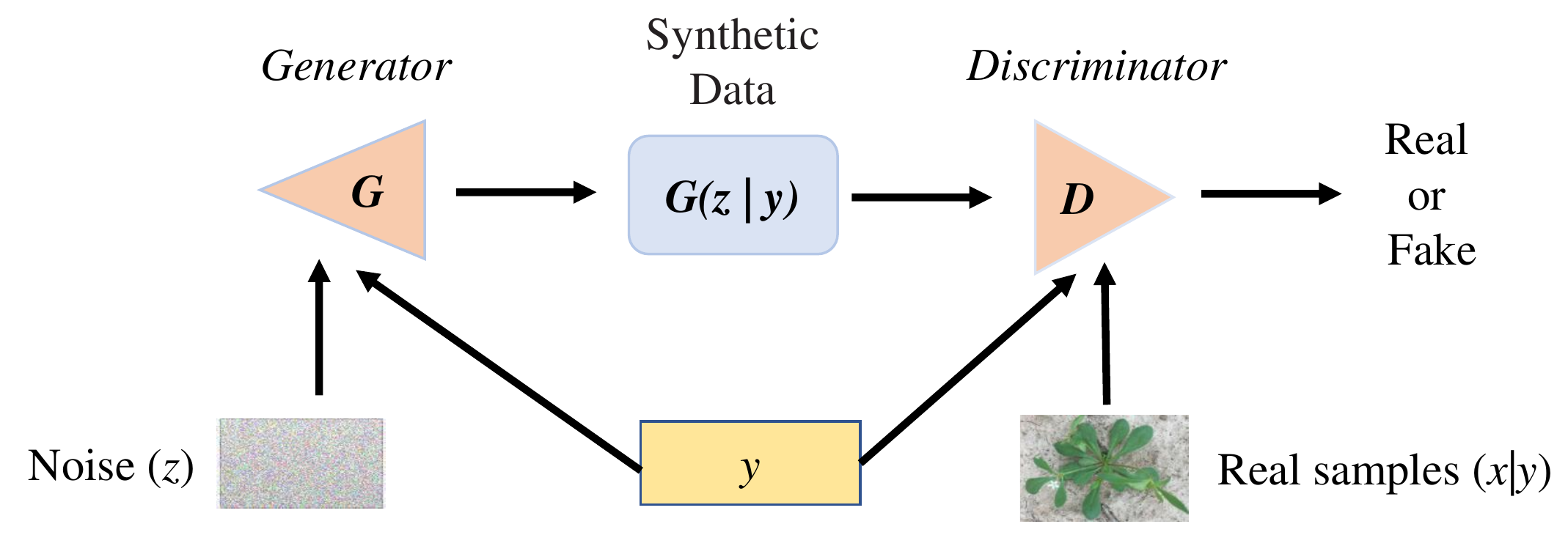}
  \caption{The framework of conditional generative adversarial network with both the \textit{generator} and \textit{discriminator} conditioned class labels ($y$).}
  \label{fig5}
\end{figure*}

\begin{figure*}[!ht]
  \centering
  \includegraphics[width=0.6\textwidth]{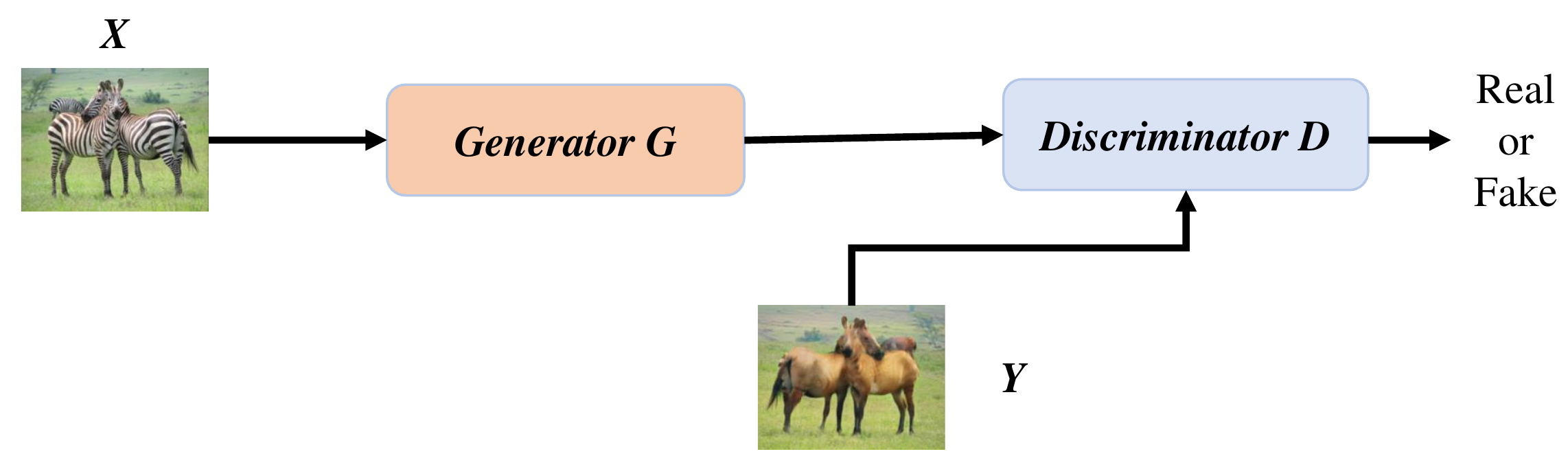}
  \caption{The framework of Pix2Pix \citep{isola2017image} for image-to-image translation where aligned or paired training data are required for training.}
  \label{fig6}
\end{figure*}

\subsection{Conditional Generative Adversarial Network}\label{sec:4.1}
The original GAN has no control over the data being generated. By conditioning the GAN on auxiliary information, the model can be enhanced to generate images with desired attributes. \cite{mirza2014conditional} who are among the first to develop conditional GANs (CGANs), conditioned both the $G$ and $D$ models on class label $y$, as illustrated in Fig.~\ref{fig5}.  The $G$ receives the input of random noise $z$ and the class label $y$ concatenated in joint representation, and the $D$ takes real samples and their corresponding class label. The objective function of CGAN is defined as follows:

\begin{equation}
    L(D,G, y) = E_{x \sim p_x}[\log D(x|y)] + E_{z \sim p_z}[\log (1-D(G(z|y)))],
\end{equation}

By \cite{mirza2014conditional}, the class-conditional GAN was able to generate MNIST digits with the class labels encoded as one-hot vectors and it was capable of learning multi-modal representations using user-generated multi-label prediction. Auxiliary classifier GAN (or AC-GAN) modifies CGAN by predicting class labels with the \textit{discriminator} instead of taking them as input \citep{odena2017conditional}, which could generate discriminable and diverse ImageNet samples exhibiting global coherence. In addition to class labels, the conditional variable $y$ can be images \citep{isola2017image, Zhu_2017_ICCV}, bounding boxes, key points, or image text descriptions. \cite{reed2016generative} generated plausible $64 \times 64$ images conditioned on text descriptions \citep{reed2016generative}.  \cite{zhang2017stackgan} proposed StackGANs for text-to-image synthesis, which could generate images of $4 \times$ better resolution. \cite{isola2017image} proposed the first general-purpose image-conditional GAN framework, called Pix2Pix (\url{https://github.com/phillipi/pix2pix}), as illustrated in Fig.~\ref{fig6}, for image-to-image translation (translating representations from source to output images). Pix2Pix requires using a training set of aligned image pairs for learning mappings between input and output images. Subsequently, \cite{Zhu_2017_ICCV} eliminated the requirement to learn mappings using unpaired training data (Section~\ref{sec:4.4}). Image-conditional GANs have been widely applied for image synthesis in face editing, image inpainting and super resolution \citep{huang2018introduction, DBLP:journals/corr/abs-2001-06937}

\subsection{Deep Convolutional Generative Adversarial Network} \label{sec:4.2}
Using convolutional layers in place of fully connected layers in the original GAN leads to the development of a family of GAN architectures known as deep convolutional GAN (DCGAN) \citep{radford2015unsupervised}, one of the most successful GAN variants widely used by many later models. In the DCGAN, convolutional layers are adopted to represent $D$ and $G$ to learn unsupervised representations, and specifically strided ($D$) and fractionally-strided ($G$) convolutions \citep{shelhamer2016fully} were utilized to learn up-sampling and down-sampling operations, which may contribute to improvements in the quality of synthesized images. To avoid model collapse (all input images mapped to the same output image, a common challenge with training GANs), in the DCGAN, batch normalization (BN) \citep{ioffe2015batch} was applied in both $D$ and $G$ during training, helping stabilize the GAN training and preventing collapsing all samples to a single sample. \cite{radford2015unsupervised} also demonstrated that proper activation functions in $G$ [rectifying linear unit (ReLU activation)] and $D$ (leaky rectified activation) can help models to learn quickly. Tested on various large-scale image datasets, DCGAN was found effective for learning the hierarchy of representations from object parts to scenes \citep{radford2015unsupervised}.

\begin{figure*}[!ht]
  \centering
  \includegraphics[width=0.6\textwidth]{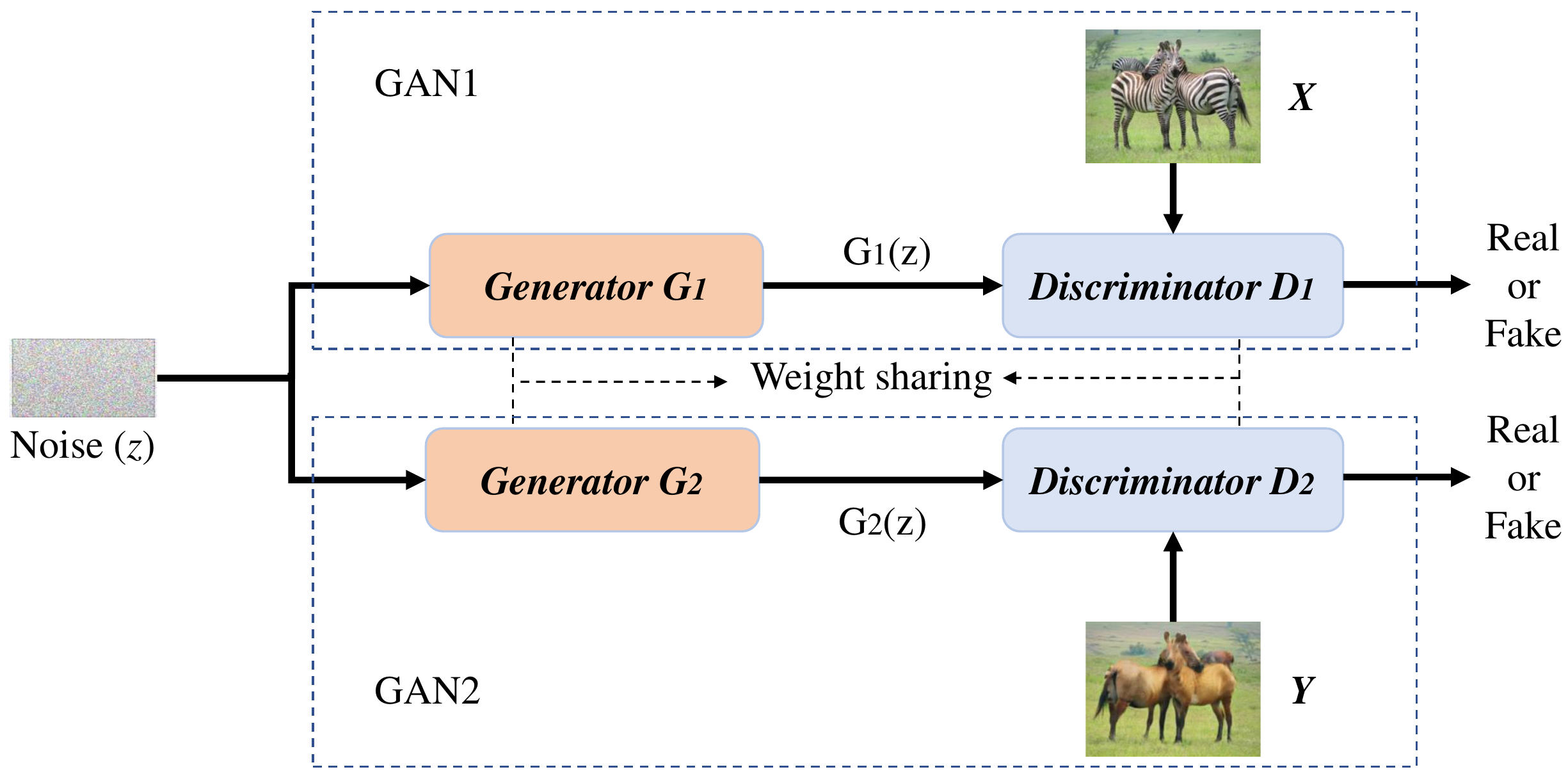}
  \caption{The framework of coupled generative adversarial network (CoGAN) for translating two image domains.}
  \label{fig7}
\end{figure*}

\subsection{Coupled Generative Adversarial Network} \label{sec:4.3}
Extending the original GAN model, coupled generative adversarial network (CoGAN) \citep{liu2016coupled} aims to learn a joint distribution of multi-domain images and generate tuples of images in different domains without correspondence supervision. The architecture of CoGAN for translating images in two domains, as illustrated in Fig.~\ref{fig7}, consists of a pair of GANs, i.e., GAN1 and GAN2, each responsible for image synthesis in one domain. During training, the two \textit{generators} ($G_1$ and $G_2$) form a team for generating a pair of images in two different domains, while the two \textit{discriminators} ($D_1$ and $D_2$) are to differentiate the images in the respective domain from those produced by the corresponding \textit{generator}. Identical network structure and weights are shared in the first layers of the \textit{generators} and in the last few layers of the \textit{discriminators} so as to enforce CoGAN to learn a joint distribution of images even without correspondence supervision between the \textit{generators} and \textit{discriminators}. The two-domain CoGAN can be generalized to multiple domains by stacking more GAN modules and enforcing the weight-sharing constraint. The CoGAN showed good performance in several joint distribution learning tasks (e.g., color and depth images) as well as image translation and domain adaptation applications ~\citep{liu2016coupled}.

\begin{figure*}[!ht]
  \centering
  \includegraphics[width=0.6\textwidth]{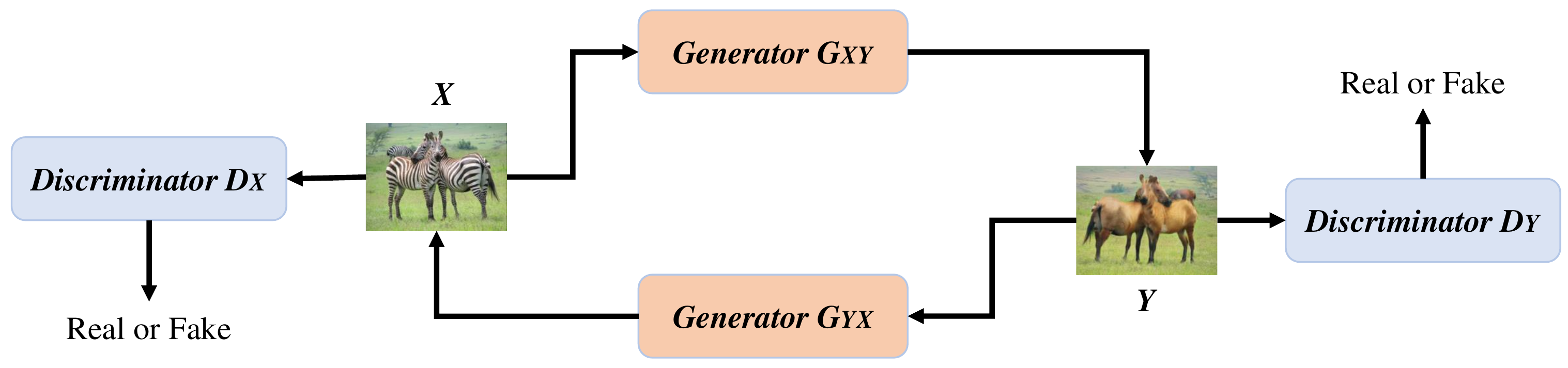}
  \caption{The framework of cycle generative adversarial network (CycleGAN) that contains four networks, i.e., two \textit{discriminators}, $D_X$ and $D_Y$, and two \textit{generator} networks, $G_{XY}$ and $G_{YX}$.}
  \label{fig8}
\end{figure*}

\subsection{Cycle-Consistent Generative Adversarial Network} \label{sec:4.4}
Built upon Pix2Pix \citep{isola2017image}, cycle-consistent generative adversarial network (CycleGAN) was designed for image-to-image translation with unpaired training data \citep{Zhu_2017_ICCV}. CycleGAN, which is trained in an unsupervised manner, eliminates the need for paired training data in Pix2Pix and is particularly useful in applications where paired training data are not readily available. CycleGAN, as shown in Fig.~\ref{fig8}, consists of two \textit{generator} networks, $G_{XY}$ and $G_{YX}$, and two \textit{discriminators}, $D_X$ and $D_Y$, where $X$ and $Y$ are source and target domains, respectively. The key idea of CycleGAN is that the two mapping functions, $G_{XY}: X \rightarrow Y$ and $G_{YX}: Y \rightarrow X$ should be inverse of each other, and the translation is to be cycle consistent (in a sense that, if we translate, e.g., zebras to horses and then translate horses back to zebras, we should get the exact same original zebra images). The adversarial loss \citep{goodfellow2014distinguishability} is applied to both mapping functions, leading to the objectives as follows:  

\begin{equation}
    \resizebox{.94\hsize}{!}{$L(D_X,G_{YX}) = E_{x \sim p_x}[\log D_x(x)] + E_{y \sim p_y}[\log (1-D_X(G_{YX}(y)))]$},
\end{equation}

\begin{equation}
    \resizebox{.94\hsize}{!}{$L(D_Y,G_{XY}) = E_{y \sim p_y}[\log D_Y(y)] + E_{x \sim p_x}[\log (1-D_Y(G_{XY}(x)))]$},
\end{equation}

To achieve cycle-consistent mappings, two $L_1$ reconstruction losses (also named cycle consistency losses) are introduced to the optimization \citep{Zhu_2017_ICCV}, with the objective function defined as follows: 
\begin{equation}
    \resizebox{.98\hsize}{!}{$L(G_{XY}, G_{YX}) = E_{x \sim p_x}[||G_{YX} (G_{XY}(x)) -x||] + E_{y \sim p_y}[||G_{XY} (G_{YX}(y)) - y||]$},
\end{equation}

Thus, the overall objective is expressed as below: 
\begin{equation}
    \resizebox{.94\hsize}{!}{$L_{CycleGAN} = L(D_x, G_{YX}) + L(D_Y, G_{XY}) + \beta L(G_{XY}, G_{YX})$},
\end{equation}
where $\beta$ is the weighting coefficient balancing the relative importance of objective functions.

The CycleGAN has achieved impressive performance in image synthesis and style transfer \citep{Zhu_2017_ICCV}. Following the idea of dual learning, CycleGAN shares structural similarities with DualGAN \citep{yi2017dualgan} for unpaired image-to-image translation, except that the former uses sigmoid cross-entropy loss format while the latter uses the loss advocated by the WGAN \citep{arjovsky2017towards}. The original CycleGAN that is unconditional has been extended to a conditional variant for image generation \citep{lu2018attribute}.

\begin{figure*}[!ht]
  \centering
  \includegraphics[width=0.6\textwidth]{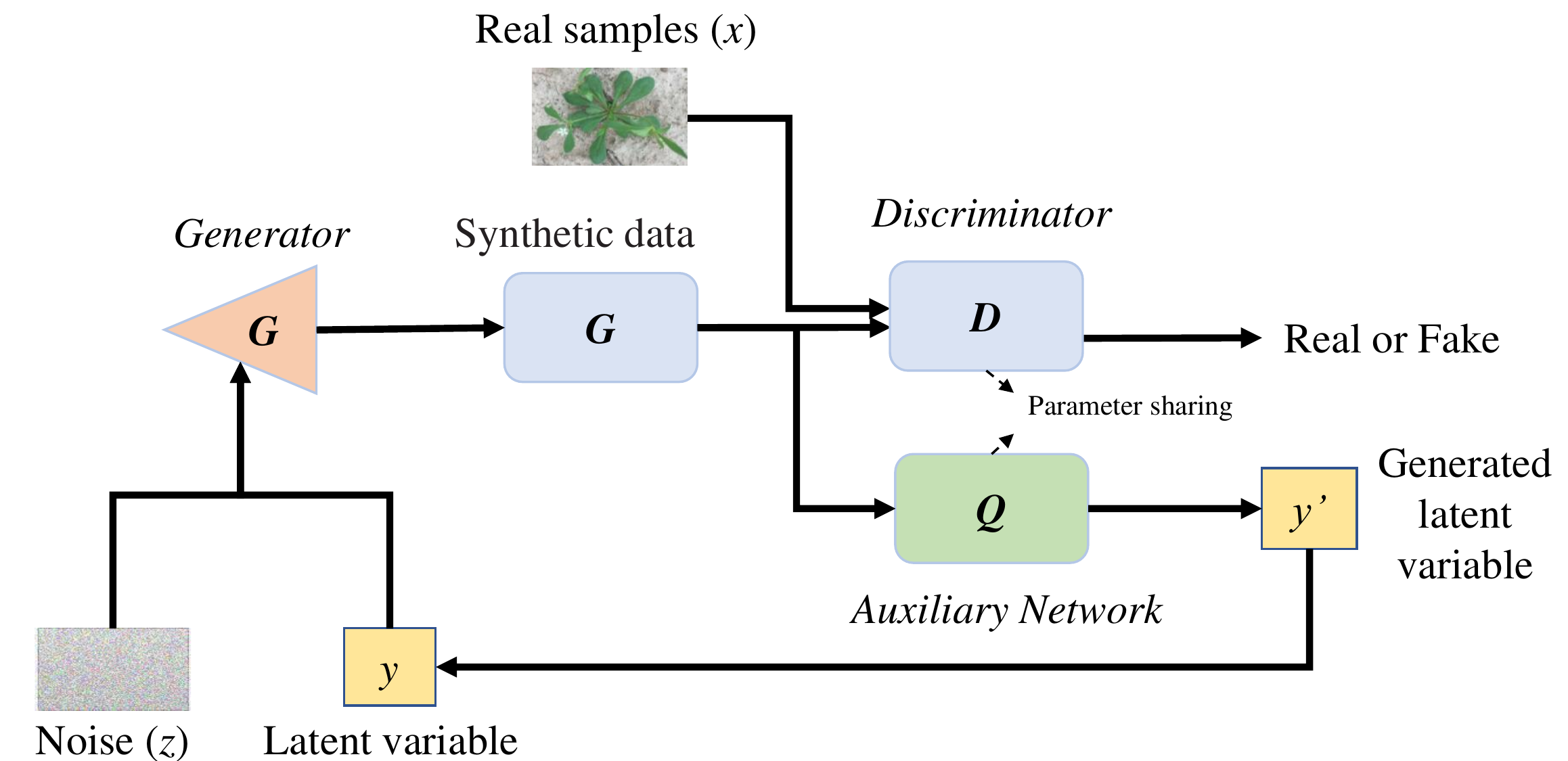}
  \caption{The framework of information maximizing generative adversarial networks.}
  \label{fig9}
\end{figure*}

\subsection{Information Maximizing Generative Adversarial Networks} \label{sec:4.5}
Information maximizing generative adversarial networks (InfoGAN), proposed by \cite{chen2016infogan}, learns the interpretable and disentangled representations in a completely unsupervised manner by maximizing the mutual information between conditional and generated data, which represents a step forward beyond the cGAN. In InfoGAN, an auxiliary network $Q$ is introduced to learn the latent variable $y'$, as shown in Fig.~\ref{fig9}, where $Q$ takes the output from $G(z|y)$ to find the representation code $y'$, while minimizing the cross entropy between $y$ and $y'$. In practice, to reduce computation costs, $Q$ and $D$ share most of the weights except the last fully connected layers. The loss function used in InfoGAN is information-regularized, which is defined as follows:
\begin{equation}
    L_{InfoGAN} = L(D, G) - \beta I(G, Q)
\end{equation}
where $L(D, G)$ is the adversarial loss in the original GAN, $I(G, Q)$ is the mutual information and $\beta$ is the tunable regularization parameter. Training InfoGAN aims to maximize the mutual information between the feature representation delivered on the input of the \textit{discriminator} and the generated image ($G(z|y)$). InfoGAN delivered impressive performance on different image datasets \citep{chen2016infogan}. There are several variants of InfoGAN, such as semi-supervision InfoGAN \citep{spurr2017guiding} and causal InfoGAN \citep{kurutach2018learning}.

\begin{figure*}[!ht]
  \centering
  \includegraphics[width=0.6\textwidth]{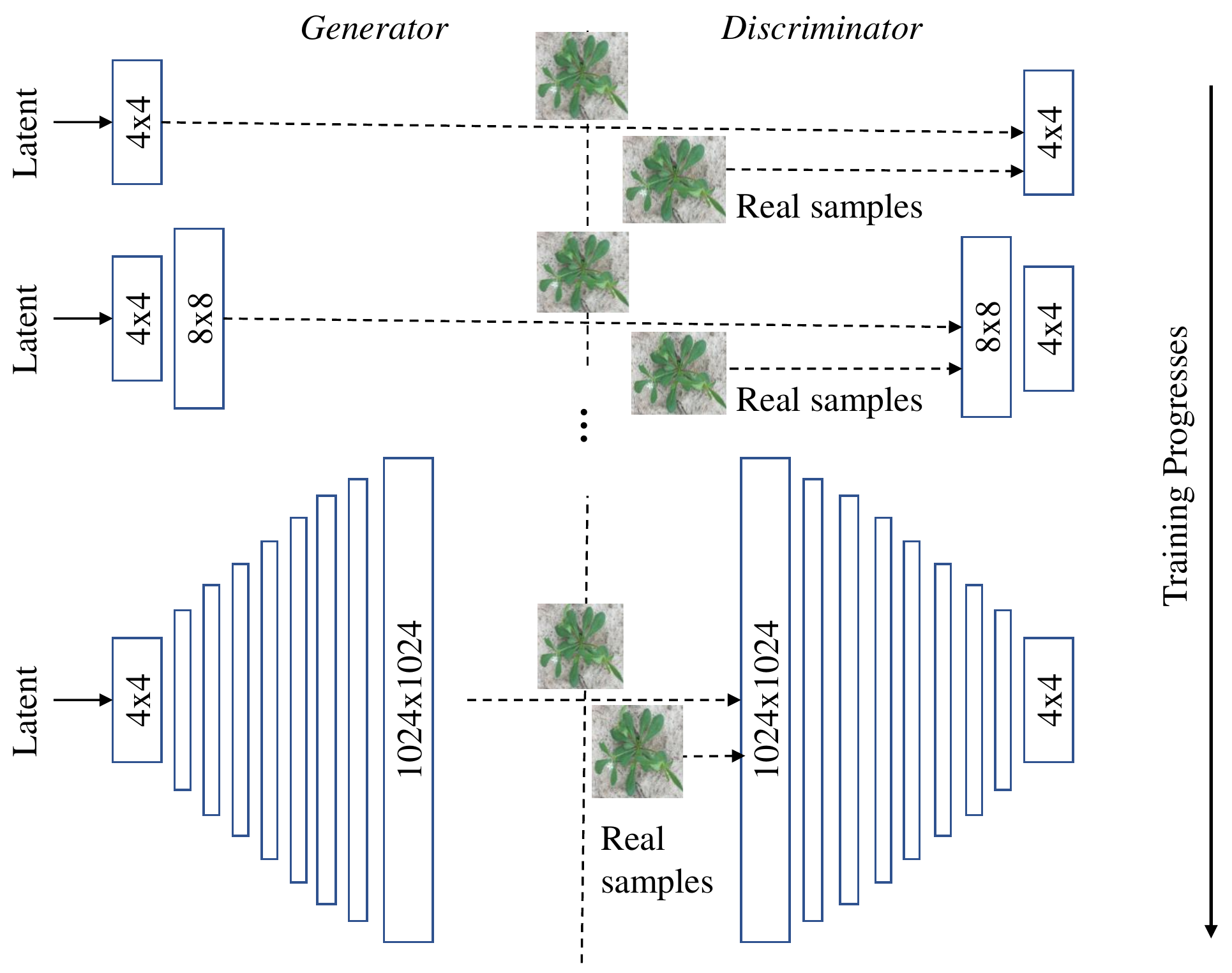}
  \caption{The architecture of progressively growing GAN (ProGAN). Adapted from \cite{karras2018progressive}.}
  \label{fig10}
\end{figure*}

\subsection{Super Resolution Generative Adversarial Network} \label{sec:4.6}
Image super-resolution aims to recover a high-resolution image upsampled from a low-resolution image, which is a fundamentally challenging task because of the ill-posed nature of the problem \citep{nasrollahi2014super}. Super resolution generative adversarial network (SR-GAN) \citep{ledig2017photo} represents the first framework to infer photo-realistic natural images for up to $4 \times$ upscaling factors by adopting the GAN framework, where the \textit{discriminator} is trained to differentiate the synthesized super-resolved images from the original images. The SR-GAN model is built with residual blocks \citep{he2016deep} as a \textit{generator} and the VGG network \citep{simonyan2014very} as a \textit{discriminator}, and it is optimized using a new perceptual loss function that is formulated as a weighted sum of adversarial loss and a content loss \citep{ledig2017photo}. The perceptual loss is defined on high-level feature maps of the VGG network that can be invariant to changes in pixel space. SR-GAN showed dramatic improvements in perceptual quality over state-of-the-art image reconstruction methods \citep{ledig2017photo}. An enhanced SR-GAN (ESR-GAN) \citep{wang2018esrgan}, significantly improves image sharpness and textural details through three major modifications made to SR-GAN. A new Residual-in-Residual Dense Block (RRDB) with BN layers removed was used as the network building block, and the Relativistic average \textit{discriminator} (RaD) \citep{jolicoeur2018relativistic} was used to predict relative realness instead of the absolute value, and the perceptual loss was improved by using features before activation instead of after activation as practiced in SR-GAN. Later, Real-ESR-GAN \citep{wang2021real}, trained with pure synthetic data, further extends ESR-GAN for restoring real-world degraded images.

\subsection{ProGAN} \label{sec:4.7}
Generating high-resolution and -fidelity images remains a challenging task for GANs since higher resolution makes it easier for the \textit{discriminator} to distinguish between generated and real samples \citep{odena2017conditional}. Progressively Growing GAN (ProGAN) \citep{karras2018progressive} employs the idea of progressive neural networks originally proposed in \cite{rusu2016progressive}. It involves progressively growing the \textit{generator} and \textit{discriminator} networks in synchrony and training the models starting from low-resolution ($4 \times 4$ pixels) incrementally to high-resolution images by adding layers to the networks, as illustrated in Fig.~\ref{fig10}. This strategy enables significantly improving training speed (2-6 times speedup) and stability at large image sizes. The incremental expansion on the convolutional layers allows the \textit{generator} and \textit{discriminator} models to effectively learn coarse-level details at the beginning and later finer-scale detail as training processes. Such a progressive training strategy is also employed by other advanced GANs such as StyleGAN \citep{karras2019style} and BigGAN \citep{brock2018large}, for synthesizing plausible images. In ProGAN, the authors also applied skip connections to add new layers either to the output of the \textit{generator} or the input of the \textit{discriminator} through a weighting scheme to control the influence of the newly added layers. During the training, an equalized learning rate was applied to all parameters, and pixel-wise feature vector normalization was done to prevent the escalation of parameter magnitudes, and WGAN-GP loss \citep{gulrajani2017improved} was used for model optimization. ProGAN was able to produce high-quality images with resolutions up to $1024 \times 1024$ pixels and achieve a record inception score (IS) \citep{salimans2016improved} of 8.8 for CIFAR10 (10 classes of $32 \times 32$ RGB images) \citep{krizhevsky2009learning}.

\begin{figure*}[!ht]
  \centering
  \includegraphics[width=0.6\textwidth]{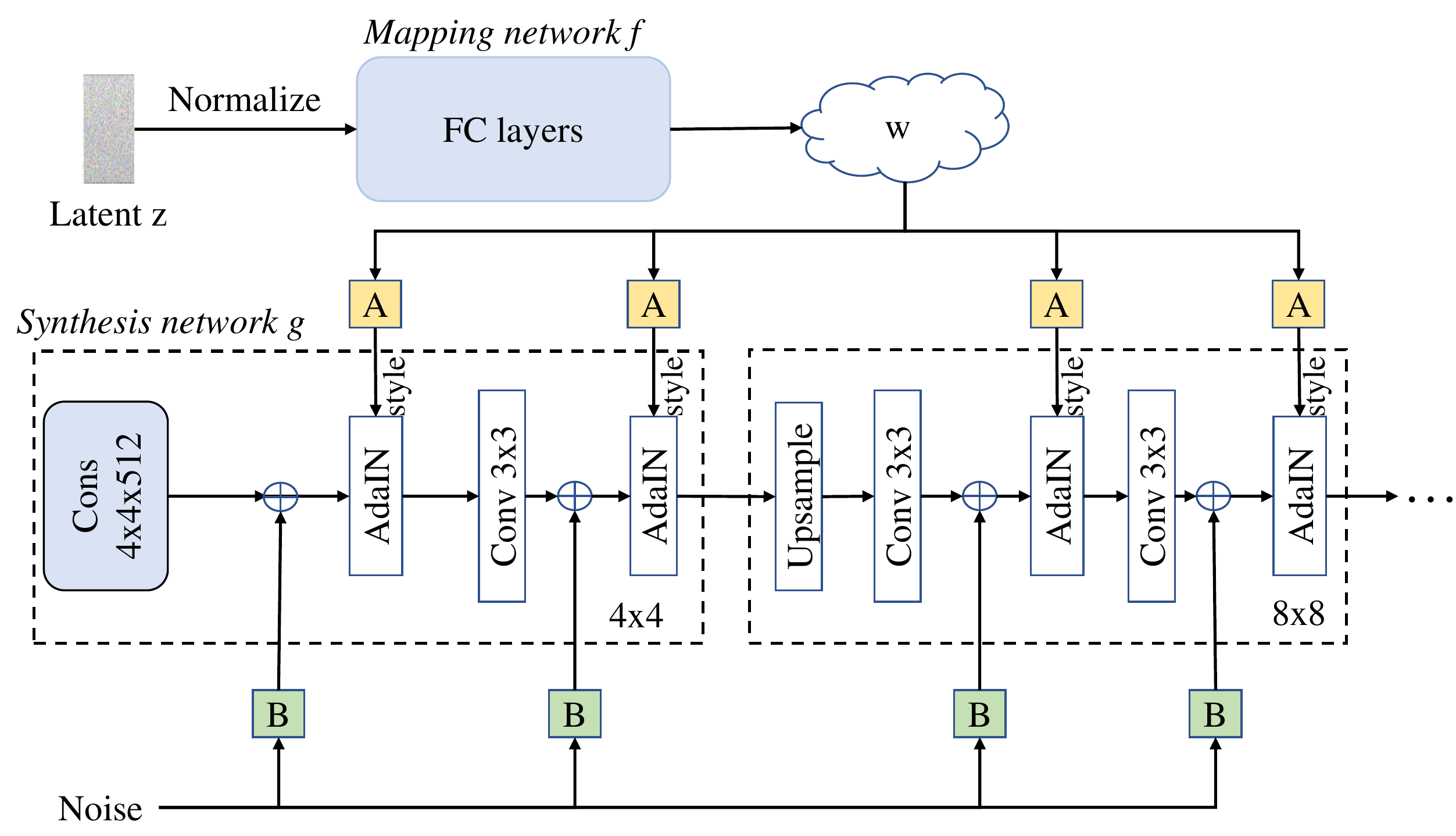}
  \caption{The architecture of style generative adversarial network. Adapted from \cite{karras2019style}}
  \label{fig11}
\end{figure*}

\subsection{Style Generative Adversarial Networks} \label{sec:4.8}
Despite developments in GAN architectures, the \textit{generator} remains a black box in image synthesis \citep{bau2018gan}.  Motivated by image style transfer \citep{huang2017arbitrary}, StyleGAN introduced by \cite{karras2019style}, is featured with a re-designed \textit{generator} architecture that enables scale-specific control over image synthesis. StyleGAN adopts the baseline of ProGAN architecture \citep{karras2018progressive} and replaces the original noise vector input of the \textit{generator} with two new references of randomness to produce a synthetic image: standalone mapping channels and noise layers as shown in Fig.~\ref{fig11}, where an adaptive instance normalization (AdaIN) layer \citep{huang2017arbitrary} is employed to control the characteristic of the generated images. 
StyleGAN employs two random latent codes in the mapping function and a split point in the synthesis network. All AdaIN operations prior to the split points use the first style latent vectors and all AdaIN operations after the split points get the second style latent vectors, which is referred to mixing regularization. This encourages the layers and blocks to localize the style to specific parts of the model and corresponding level of detail in the generated image. Stochastic variation is introduced through adding Gaussian noise in the \textit{generator} model and broadcasted to all feature maps, which enables the generation network to synthesize images diverse in a fine-grained, per-pixel manner. In \cite{karras2019style}, StyleGAN achieved about 20\% improvements over the ProGAN in terms of Fréchet inception distance (FID) \citep{heusel2017gans}. Built upon the StyleGAN, two improved versions, StyleGAN2 \citep{karras2020analyzing} and StyleGAN3 \citep{karras2021alias}, have been developed to remove image artifacts (e.g., texture sticking, unwanted information leaked into the image synthesis process) for better quality of generated images.

\begin{figure*}[!ht]
  \centering
  \includegraphics[width=0.6\textwidth]{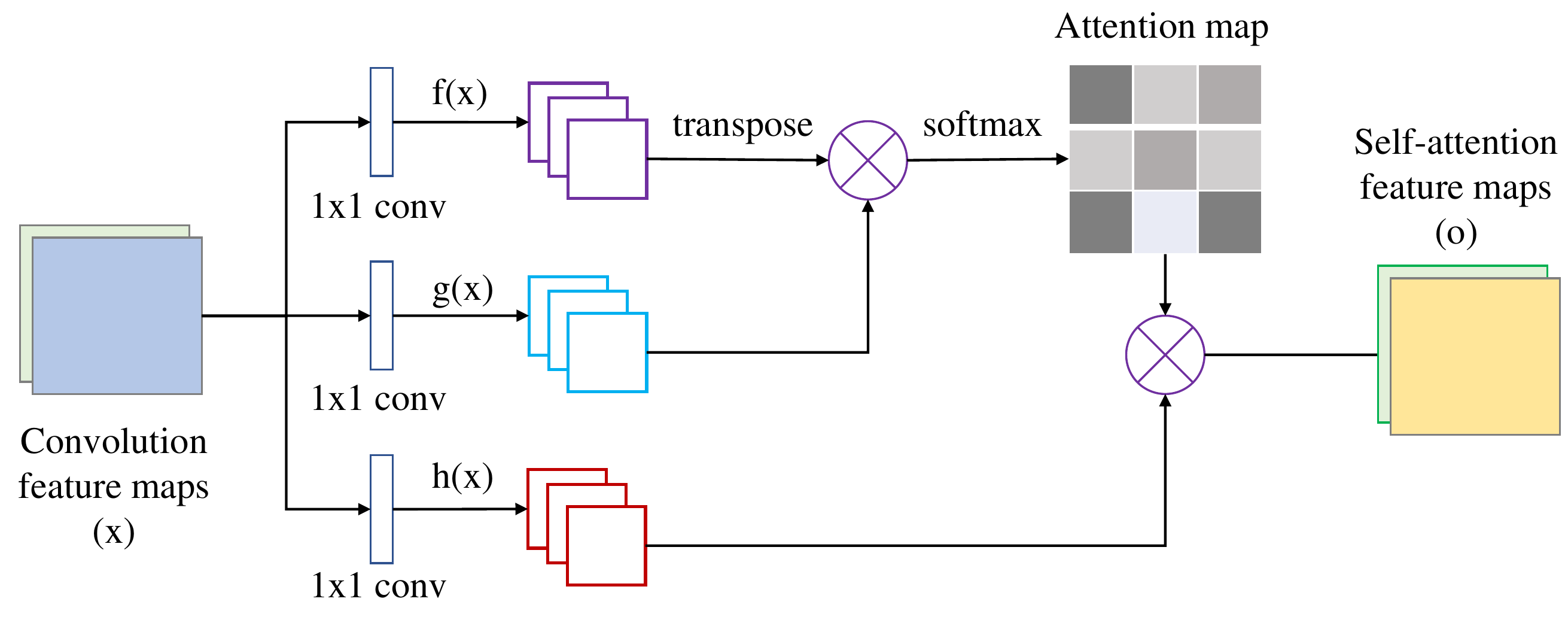}
  \caption{The architecture of the self-attention network used in self-attention generative adversarial network. Adapted from \cite{zhang2019self}.}
  \label{fig12}
\end{figure*}

\subsection{Self-Attention Generative Adversarial Network} \label{sec:4.9}
CNNs are commonly used to build GANs for image generation. Since CNNs generally process information in a local neighborhood, convolutional GANs have difficulty in modeling long-range dependencies beyond local regions. They may fail to capture geometric or structure patterns that occur consistently in some classes when learning multi-class image datasets. \cite{zhang2019self} proposed the Self-Attention GAN (SA-GAN) for modeling long-range dependencies for image generation tasks. SA-GAN incorporates a self-attention mechanism \citep{vaswani2017attention}, as shown in Fig.~\ref{fig12}, into the convolutional GAN framework. Adapted from the non-local model of \cite{wang2018non}, the self-attention module is deployed in the design of the \textit{generator} and \textit{discriminator} architectures \citep{zhang2019self}, which is complementary to convolutions for image generation. SA-GAN enables a large receptive field for CNNs to model long-range, multi-level dependencies across image regions, without sacrificing computational efficiency. In SA-GAN, spectral normalization (SN) \citep{miyato2018spectral} is applied to the \textit{generator} for stabilized training. Experimented on the ImageNet dataset, SA-GAN achieved state-of-the-art performance on multi-class image synthesis \citep{zhang2019self}. 

\begin{figure*}[!ht]
  \centering
  \includegraphics[width=0.75\textwidth]{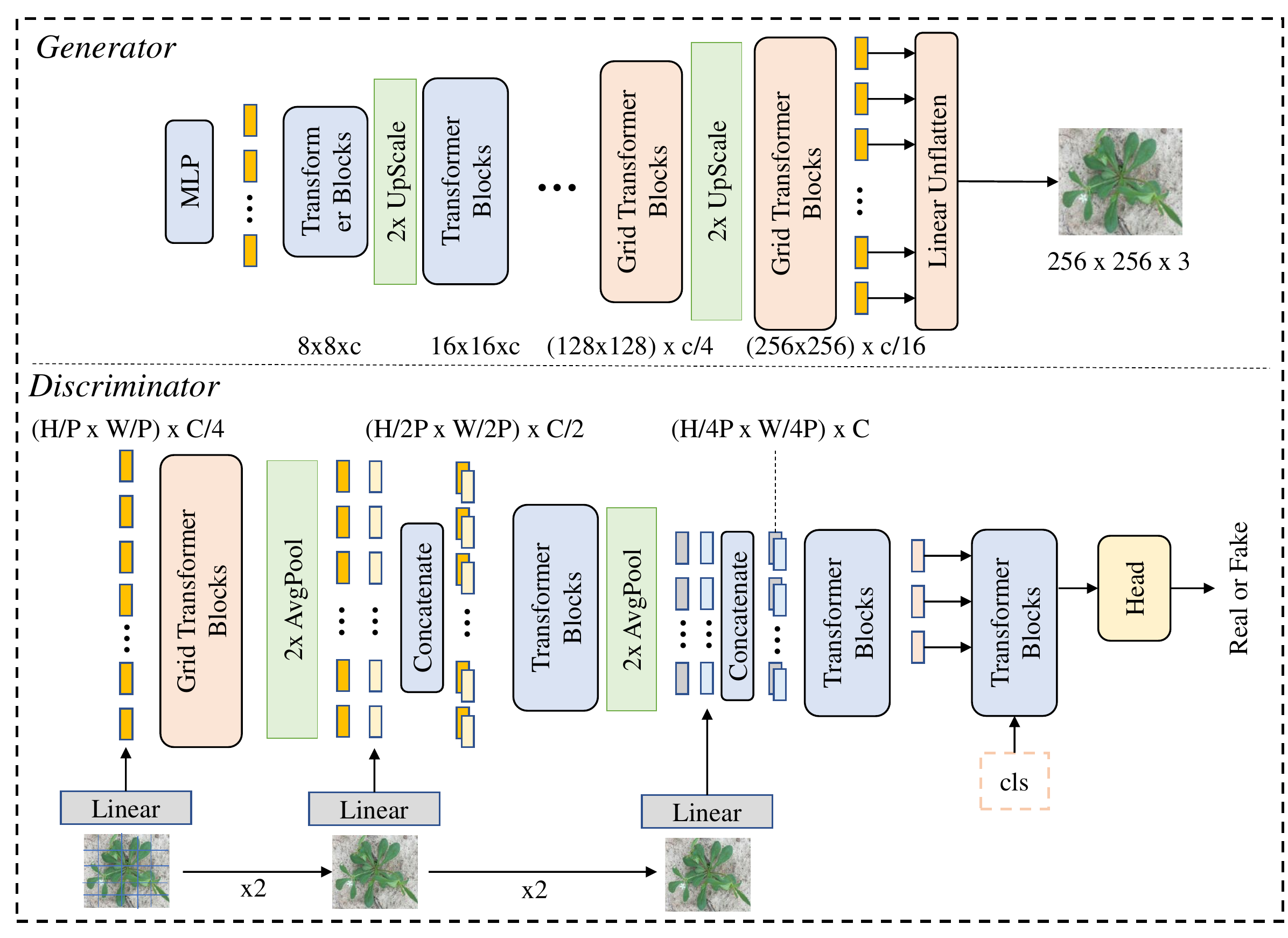}
  \caption{The framework of a transformer generative adversarial network. Adapted from \cite{wang2021lychee}.}
  \label{fig13}
\end{figure*}

\subsection{BigGAN} \label{sec:4.10}
BigGAN \citep{brock2018large} was designed for scaling up GAN models for generating high-resolution, high-fidelity images through a regularization scheme. Built on the SA-GAN architecture \citep{zhang2019self}, BigGAN employs a set of regularization techniques for scaling-up image synthesis, such as moving average of \textit{generator}’s weights, class-conditional BatchNorm \citep{dumoulin2016learned}, orthogonal initialization scheme, direct skip connections (skip-z), larger model and batch size, truncation trick, etc. Particularly, BigGAN adopts a shared embedding approach in the BatchNorm layers to significantly reduce computation costs and improve training speed. The authors successfully trained the model on ImageNet images at $128 \times 128$, $256 \times 256$ and $512 \times 512$ resolutions \citep{brock2018large}, improving on the state of the art by a large margin. It is demonstrated that GANs benefit from larger models (4 times bigger) and larger batch sizes (8 times larger) since large networks and batch size can significantly increase the capacity of models to handle complex datasets like ImageNet. 

\begin{table*}[!ht]
\renewcommand{\arraystretch}{1.6}
\caption{Summary of GAN variants with open-source software packages.}
\centering
\label{tab2}
\resizebox{0.8\textwidth}{!}{
\begin{tabular}{l|l|l}
\hline
GAN variants & URL                                                  & Reference                  \\ \hline
AutoGAN      & \url{https://github.com/VITA-Group/AutoGAN}                & \cite{gong2019autogan}         \\ \hline
BigGAN       & \url{https://github.com/ajbrock/BigGAN-PyTorch}            & \cite{brock2018large}        \\ \hline
CGAN/CDCGAN & \url{https://github.com/znxlwm/tensorflow-MNIST-cGAN-cDCGAN}                                    &\cite{mirza2014conditional} \\ \hline
CoGAN        & \url{https://github.com/mingyuliutw/cogan}                 & \cite{liu2016coupled}        \\ \hline
CycleGAN    & \url{https://github.com/junyanz/pytorch-CycleGAN-and-pix2pix}                                   & \cite{Zhu_2017_ICCV}          \\ \hline
DCGAN       & \url{https://github.com/eriklindernoren/PyTorch-GAN/tree/master/implementations/dcgan} & \cite{radford2015unsupervised}      \\ \hline
ESR-GAN      & \url{https://github.com/xinntao/ESRGAN }                   & \cite{wang2018esrgan}         \\ \hline
InfoGAN      & \url{https://github.com/openai/InfoGAN}                    & \cite{chen2016infogan}         \\ \hline
Vanilla GAN  & \url{https://github.com/goodfeli/adversarial}              & \cite{goodfellow2014generative} \\ \hline
Pix2Pix      & \url{https://github.com/phillipi/pix2pix}                  & \cite{isola2017image}         \\ \hline
ProGAN      & \url{https://github.com/tkarras/progressive\_growing\_of\_gans}                                 & \cite{karras2018progressive}       \\ \hline
SA-GAN       & \url{https://github.com/brain-research/self-attention-gan} & \cite{zhang2019self}.         \\ \hline
StyleGAN     & \url{https://github.com/NVlabs/stylegan}                   & \cite{karras2019style}       \\ \hline
StyleGAN2    & \url{https://github.com/NVlabs/stylegan2}                  & \cite{karras2020analyzing}       \\ \hline
StyleGAN3    & \url{https://github.com/NVlabs/stylegan3}                  & \cite{karras2021alias}       \\ \hline
StyleSwin    & \url{https://github.com/microsoft/StyleSwin}               & \cite{zhang2021styleswin}        \\ \hline
SR-GAN       & \url{https://github.com/tensorlayer/srgan}                 & \cite{ledig2017photo}        \\ \hline
TransGAN     & \url{https://github.com/asarigun/TransGAN}                 & \cite{jiang2021transgan}        \\ \hline
WGAN         & \url{https://github.com/eriklindernoren/Keras-GAN}         & \cite{arjovsky2017wasserstein}     \\ \hline
WGAN-GP      & \url{https://github.com/caogang/wgan-gp}                   & \cite{gulrajani2017improved}    \\ \hline
\end{tabular}
}
\end{table*}

\subsection{Transformer Generative Adversarial Network} \label{sec:4.11}
While widely used in modern GAN architectures, CNNs are considered not well suited to process long-range dependencies because of limited local receptive fields, which may lead to loss of feature resolution and fine details as well as the difficulty of model optimization. To address these potential issues, \cite{jiang2021transgan} proposed a novel transformer-based GAN architecture completely free of convolutions, called TransGAN. Transformer \citep{han2022pyramidtnt} is an emerging family of DL methods based on self-attention mechanisms, initially applied to natural language processing and subsequently spread in computer vision applications. Transformers enable modeling long dependencies between input sequence elements and support parallel processing of sequence, which otherwise would not be readily achieved by CNNs. In TransGAN, the pure transformer-based \textit{generator} and \textit{discriminator} are used in place of CNN as GAN building blocks. For memory efficiency, the \textit{generator} consists of multiple transformer modules, which progressively increases the feature map resolution, as shown in Fig.~\ref{fig13}. Furthermore, a multi-scale structure \textit{discriminator} is used to capture semantic contexts and low-level textures simultaneously. Training techniques including data augmentation and modified layer normalization are adopted to stabilize model optimization and generalization. Tested on low-resolution (e.g., $48 \times 48$ and $128 \times 128$) benchmark datasets, TransGAN outperformed several state-of-the-art CNN-based GANs \citep{jiang2021transgan}, such as WGAN-GP \citep{gulrajani2017improved}, SN-GAN \citep{miyato2018spectral}, AutoGAN \citep{gong2019autogan} and StyleGAN-V2 \citep{karras2020analyzing}. However, the performance of TransGANs in synthesizing high-resolution images (e.g., $1024 \times 1024$) remains to be improved or extensively assessed. 

Very recently, \cite{zhang2021styleswin} proposed GAN called StyleSwin that adopts Swin transformer in a style-based architecture for high-resolution image generation. StyleSwin is featured with three core architectural adaptations, i.e., a local attention in the style-based architecture, double attention, and sinusoidal positional encoding. It outperformed prior transformer-based GNAs and StyleGAN and achieved comparable performance as StyleGAN2 \citep{zhang2021styleswin}.

\subsection{Summary} \label{sec:4.12}
Table~\ref{tab2}  summarizes various GAN architectures developed for image synthesis. These architectures do not represent the full landscape of GANs in literature, but they have open-sourced packages and have been employed or adapted for expanding agricultural image data, as reviewed in Section~\ref{sec:app_gans}. In addition to the tabular summary, readers are referred to dedicated literature \citep{wang2021lychee,huang2018introduction, creswell2018generative} and ``The GAN Zoo'' in the Github (\url{https://github.com/hindupuravinash/the-gan-zoo}) for a more comprehensive list of GAN variants as well as a GAN toolbox in the PyTorch environment at \url{https://github.com/facebookresearch/pytorch_GAN_zoo} for algorithm implementation. 

\section{Applications of GAN in Agriculture}
\label{sec:app_gans}
Presented in this section are a summary review of application studies of GAN in the agricultural field and food domain. These applications are loosely organized into three areas including precision agriculture (Table~\ref{tab3}), plant phenotyping (Table~\ref{tab4}), and postharvest quality assessment of agricultural products (Table~\ref{tab5}).  

\subsection{Precision Agriculture} \label{sec:5.1}
Precision agriculture, also known as precision or smart farming, aims to enhance crop and animal production using more precise (e.g., site-specific) and resource-efficient approaches \citep{monteiro2021precision}. It exploits an assortment of monitoring and invention technologies (e.g., imaging, autonomy, AI/robotics) to support data-driven agricultural tasks, such as crop health detection, weed recognition and control, and precision livestock farming. Over the past few years, GANs have been increasingly applied to augment image data in precision agriculture and enhance machine learning models for various applications as reviewed below.

\subsubsection{Plant Health} \label{sec:5.1.1}
Biotic and abiotic stressors \citep{dhaka2021survey}, such as microorganisms (e.g., virus, bacteria, fungi), insects, and environmental factors, negatively affect plant growth and health conditions, leading to the development of plant diseases or disorders and eventually low yield and quality of plant products \citep{zhang2021identification}. Imaging technologies (e.g., RGB, multi-/-hyper-spectral, fluorescence, thermal) offer a non-invasive and objective means for characterization and diagnosis of plant health conditions (e.g., plant disease detection) \citep{thomas2018benefits,mahlein2018hyperspectral}. One of the grand challenges in imaging-based plant health detection is how to cope with limited image data with a small number of expert annotations, especially for emerging plant diseases for which it can be very difficult to accumulate large amounts of image data ground-truthed with specialized knowledge. This in turn hampers the implementation of supervised machine learning methods (e.g., CNNs). The developments of GANs for image synthesis, given successful applications in medical diagnosis \citep{yi2019generative}, has inspired significant efforts made to utilize GANs to augment image data for enhancing the detection of plant diseases and other health conditions.  

CGAN (Section~\ref{sec:4.1})/DCGAN (Section~\ref{sec:4.2}) have received a great deal of attention in image augmentation for improved plant disease recognition. \cite{hu2019low} reported on using DCGAN with conditional label constraint (named C-DCGAN) for identifying tea leaf diseases (red scab, red leaf spot and leaf blight). Original color images, subjected to disease segmentation by support vector machine (SVM), were augmented from 20 to 4980 images per disease and fed into classifiers (VGG16, SVM, decision tree and random forest). With the GAN-augmented data, VGG16 yielded the best average accuracy of 90\%, which was 28\% higher than that using rotation and translation-based augmentation and even higher than the accuracy by VGG16 without data augmentation due to severe overfitting.  \cite{abbas2021tomato} used CGAN (Section~\ref{sec:4.1}) to augment tomato leaf images of ten classes, curated from the Plant Village dataset \citep{hughes2015open}. With 4000 synthetic images per class added to training data, classification accuracies of over 97\% were achieved by DenseNet121, which were about 1-4\% higher than the accuracies on the original data without image augmentation. 


\cite{douarre2019novel} presented an early study using DCGAN to generate images for segmenting apple scab (a disease manifested as spot lesions on the leaves and fruits). The infrared (IR) images acquired for plant canopy were titled into sub-images of $64 \times 64$ pixels and fed into SegNet \citep{badrinarayanan2017segnet} for pixel-wise segmentation (“scab” vs “not scab”). Compared to a baseline model without involving data augmentation, DCGAN with the WGAN training process \citep{arjovsky2017towards} yielded a 2\% improvement in the segmentation accuracy. Larger improvements were, however, achieved using standard data augmentation and model-based simulation approach \citep{douarre2019novel}, because DCGAN was found difficult to converge to produce convincing images.  \cite{zeng2020gans} curated a citrus leaf dataset comprising 5406 color images with visual symptoms of Huanglongbing (a citrus greening disease) from the Plant Village dataset \citep{hughes2015open}. DCGAN was adapted to expand the training data (80\% of the total), synthesizing additional 8650 images (doubling the number of training samples). An improvement of up to 20\% classification accuracy was achieved using the GAN-augmented data.  \cite{zhang2019classification} also employed DCGAN to expand images of citrus canker disease and obtained a classification accuracy of 90.1\%. A further study was conducted on generating lesion images with a specific shape by feeding binary images into the GAN \textit{generator} \citep{sun2020data}, which improved the recognition of leaf lesion from 95.5\% on the original data to 97.8\% for synthetic images.

\cite{yuwana2020data} investigated the vanilla GAN (with multilayer perceptron as \textit{generator} and \textit{discriminator} models) and DCGAN to synthesize tea leaf images of four classes (healthy and three diseased) for disease identification. With 1000 images synthesized per class, the vanilla GAN and DCGAN yielded classification accuracies of 88.84\% and 88.86\%, respectively, representing about 2.5\% improvements over the baseline models without image augmentation. However, generating more training samples led to mixed performance, probably due to the inclusion of more low-quality (e.g., noisy, texture detail loss) images generated by the GANs. \cite{wu2020dcgan} applied DCGAN to generate images of tomato leaves of 5 classes (healthy and four diseased). A set of 1500 images extracted from the Plant Village dataset \citep{hughes2015open} were augmented to 5300 images to train CNN classification models, resulting in accuracies of 79\%-94\%. However, the authors did not systematically compare models trained with and without using DCGAN. In classifying images of seven tomato leaf diseases, \cite{hu2021edge} used DCGAN to augment image data for different CNN classifiers (VGG19, AlexNet and ResNet50) and achieved accuracy improvements of 1\% to 3\%, compared to models without image augmentation. In using DCGAN to generate images for early identification of tomato plants infected tomato mosaic virus, \cite{gomaa2021early} obtained the recognition accuracy of 97.9\% using the augmented data, about one percent better than the accuracy without augmentation.

CycleGAN (see Section~\ref{sec:4.4}) has also been evaluated to synthesize and augment image data in plant health detection. \cite{nazki2019image} applied CycleGAN with U-Net \citep{ronneberger2015u} as a \textit{generator} for image synthesis to rebalance a 9-class tomato disease dataset. In terms of perceptual image quality metrics such as FID, CycleGAN with U-Net performed better than the classic CycleGAN in capturing low-level details and realistic texture. However, this study did not examine the performance of synthesized images in plant disease recognition tasks. In applying CycleGAN to a small 4-class grape disease dataset, \cite{zeng2021few} obtained accuracy improvements of over 7\% with ResNet18, compared to the model on the original data. \cite{tian2019detection} used CycleGAN to complement traditional data augmentation methods for enhanced detection of anthracnose lesions on apple fruits. The dataset, collected from both orchards and online, consisted of 140 diseased and 500 healthy apple images, and CycleGAN was applied to transform healthy apple images into diseased fruit images. With the inclusion of CycleGAN-synthesized images, the disease detection model, based on YOLO-V3 \citep{redmon2018yolov3}, gained about 5\% improvements in terms of F1 score and IoU (intersection over union). Despite these applications, CycleGAN has limitations in generating high-quality images because of no explicit attention mechanisms for transforming specific objects in images, and consequently the generated images may contribute little to model robustness in plant disease diagnosis \citep{cap2020leafgan}. To enrich the versatility of image generation, \cite{cap2020leafgan} introduced a leaf segmentation module (composed of a weakly supervised segmentation network) to CycleGAN, leading to a model called LeafGAN, to transform regions of interest in plant disease images. Tested on 5-class cucumber leaf disease data, LeafGAN yielded diagnostic performance improvements by 7.4\% as opposed to 0.7\% by the vanilla CycleGAN.

\begin{figure*}[!ht]
  \centering
  \includegraphics[width=0.5\textwidth]{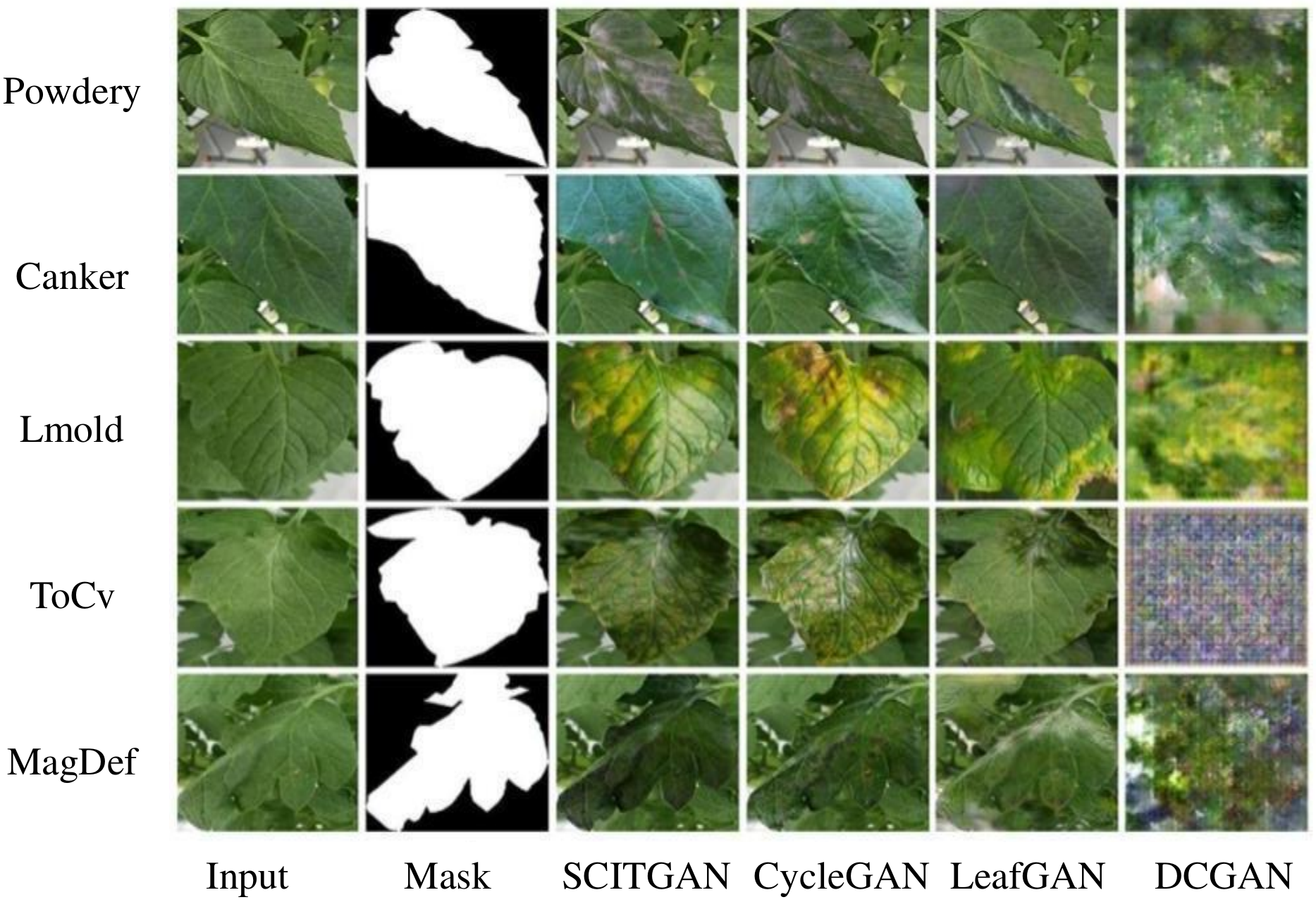}
  \caption{Sample examples generated by different GAN models (CycleGAN, LeafGAN, DCGAN) against the style consistent image translation GAN \citep{xu2021style}.}
  \label{fig15}
\end{figure*}

High-resolution images are crucial for effective recognition and detection of plant diseases. SR-GAN (Section~\ref{sec:4.6}) enables synthesizing super-/high-resolution images with photo-realistic details from down-sampled, low-resolution images. \cite{maqsood2021super} examined SR-GAN combined with a CNN model for classifying wheat yellow rust disease images. The original data, consisting of 1922 color images of three classes (healthy, resistant, and susceptible) were first preprocessed to prepare down-sampled, low-resolution images, and then fed into SR-GAN with an up-sampling factor of $4 \times$ to produce photorealistic images. The synthetic images yielded an overall classification accuracy of 83\%, which outperforms the down-sampled, low-resolution images that only gave 75\% overall accuracy. However, this study did not apply SR-GAN to the original images for image synthesis. Built upon SR-GAN, an enhanced version, i.e., ESR-GAN \citep{wang2018esrgan}, can produce better visual quality with more realistic textures. \cite{wen2020crop} reported on using ESR-GAN for image synthesis for a tomato disease dataset extracted from the Plant Village dataset \citep{hughes2015open}.Likewise, ESR-GAN was applied to low-resolution target images with a scaling factor of $\times 4$ and fine-tuned for the domain task. The generated images, trained on VGG16, yielded an accuracy of 90.78\% in the 10-class classification, substantially better than 72.74\% by using low-resolution images. It is noted that the accuracy obtained from the synthesized SR images is still about 4\% lower than that by the original high-resolution images. 

StyleGAN (see Section~\ref{sec:4.8}) that combines ProGAN and neural style transfer has also been used to generate high-resolution images for enhanced plant disease recognition. In \cite{arsenovic2019solving}, StyleGAN along with traditional data augmentation methods (geometric transformations, such as rotations or pixel-wise changes) was employed to augment the Plant Disease dataset \citep{brahimi2017deep}, containing 79265 images of 12 different species and 42 classes (including both healthy and diseased). To compare the approaches while considering the impact of environment conditions, the authors trained five CNN plant disease classification models on the PlantVillage dataset \citep{hughes2015open} and tested the models on the Plant Disease dataset \citep{brahimi2017deep} with both standard augmentation and the StyleGAN. The models trained on the StyleGAN-augmented dataset (top-1 accuracy of 0.9088 for ResNet-152) outperformed the models trained on the dataset with standard augmentation (top-1 accuracy of 0.8995 for ResNet-152). The authors also reported promising performance for leave detection algorithms, but performance comparisons regarding data augmentation were not detailed.

Apart from the aforementioned GAN methods, some other GAN variants have also been used for plant health detection. \cite{lu2019generative} reported on using AC-GAN (auxiliary classifier GAN) \citep{odena2017conditional} to generate synthetic insect pest images for classifying 5-class insect images and obtained an F1-score of 0.95 (0.3 higher than that based on traditional data augmentation). \cite{nazki2020unsupervised} proposed AR-GAN (activation reconstruction GAN) by introducing a module to calculate activation reconstruction loss in CycleGAN \citep{Zhu_2017_ICCV} to rebalance an imbalanced 9-class tomato disease dataset. The GAN-synthesized data yielded an improvement of 5.2\% in classification accuracy as compared to an 0.8\% increase with traditional augmentation. The AR-GAN approach was later adopted by \cite{zhang2021identification} for improving the identification of cucumber leaf diseases. \cite{liu2020data} presented a GAN model with a channel decreasing \textit{generator} to synthesize 4-class grape leaf images, reporting 98.7\% classification accuracy, which is about 3\% and 2\% better than the models without and with only basic image augmentation, respectively. Recently, \cite{xu2021style} adapted style consistent image translation GAN (SCITGAN) \citep{wang2019example} for tomato disease recognition. A five-class tomato leaf dataset was collected for GAN training to generate images (Fig.~\ref{fig15}) and instances for each disease class. Among different object detection and instance segmentation models, PointRend \citep{kirillov2020pointrend} trained with the SCITGAN-augmented data achieved the best segmentation accuracy with mAP (mean average precision) of 68.3\%, outperforming the models with CycleGAN (67.6\%) and without augmentation (56.1\%). More application studies of GANs on plant health detection can be found in Table~\ref{tab3}.

\begin{table*}[!ht]
\renewcommand{\arraystretch}{1.4}
\centering
\caption{Various GAN methods applied for plant health  detection.}
\label{tab3}
\resizebox{0.65\textwidth}{!}{
\begin{tabular}{ll|l|ll}
\hline
\multicolumn{2}{l|}{GAN   Architecture}           & Plant   Materials      & \multicolumn{2}{l}{Reference(s)}                           \\ \hline
\multicolumn{2}{l|}{\multirow{2}{*}{CGAN/CDCGAN}} & Tea leaves             & \multicolumn{2}{l}{\cite{hu2019low}}                       \\ \cline{3-5} 
\multicolumn{2}{l|}{}                             & Tomato leaves          & \multicolumn{2}{l}{\cite{abbas2021tomato}}                    \\ \hline
\multicolumn{2}{l|}{\multirow{6}{*}{DCGAN}}       & Apple scab             & \multicolumn{2}{l}{\cite{douarre2019novel}}                  \\ \cline{3-5} 
\multicolumn{2}{l|}{}                             & Tomato leaves          & \multicolumn{2}{l}{\cite{wu2020dcgan,hu2021edge}}    \\ \cline{3-5} 
\multicolumn{2}{l|}{}                             & Citrus canker          & \multicolumn{2}{l}{\cite{zhang2019classification,sun2020data}} \\ \cline{3-5} 
\multicolumn{2}{l|}{}                             & Citrus leaves          & \multicolumn{2}{l}{\cite{zeng2020gans}}                     \\ \cline{3-5} 
\multicolumn{2}{l|}{}                             & Tea leaves             & \multicolumn{2}{l}{\cite{yuwana2020data})}                   \\ \cline{3-5} 
\multicolumn{2}{l|}{}                             & Blueberry leaves       & \multicolumn{2}{l}{\cite{kim2021fruit}}                      \\ \hline
\multicolumn{2}{l|}{\multirow{4}{*}{CycleGAN}}                                                 & Tomato leaves          & \multicolumn{2}{l}{\cite{nazki2019image}}     \\ \cline{3-5} 
\multicolumn{2}{l|}{}                             & Apple leaves          & \multicolumn{2}{l}{\cite{tian2019detection}}                     \\ \cline{3-5} 
\multicolumn{2}{l|}{}                             & Cucumber leaves        & \multicolumn{2}{l}{\cite{cap2020leafgan}}                      \\ \cline{3-5} 
\multicolumn{2}{l|}{}                             & Grape leaves           & \multicolumn{2}{l}{\cite{zeng2021few}}                     \\ \hline
\multicolumn{2}{l|}{Info-GAN}                     & Tomato leaves         & \multicolumn{2}{l}{\cite{gomaa2021early}}               \\ \hline
\multicolumn{2}{l|}{SR- GAN}                      & Wheat stripe rust      & \multicolumn{2}{l}{\cite{maqsood2021super}}                  \\ \hline
\multicolumn{2}{l|}{ESR-GAN}                      & Tomato leaves          & \multicolumn{2}{l}{\cite{wen2020crop}}                      \\ \hline
\multicolumn{2}{l|}{StyleGAN}                     & Multiple species       & \multicolumn{2}{l}{\cite{arsenovic2019solving}}                \\ \hline
\multicolumn{1}{l|}{\multirow{11}{*}{Others}} & AC-GAN                                         & Insect pest            & \multicolumn{2}{l}{\cite{lu2019generative}}         \\ \cline{2-5} 
\multicolumn{1}{l|}{}                         & \multirow{2}{*}{Activation reconstruction GAN} & Tomato leaves          & \multicolumn{2}{l}{\cite{nazki2020unsupervised}}      \\ \cline{3-5} 
\multicolumn{1}{l|}{}   &                         & Cucumber leaves        & \multicolumn{2}{l}{\cite{zhang2021identification}}                    \\ \cline{2-5} 
\multicolumn{1}{l|}{}   & Channel decreasing GAN  & Grape leaves           & \multicolumn{2}{l}{\cite{liu2020data}}                      \\ \cline{2-5} 
\multicolumn{1}{l|}{}   & PSR-GAN                 & Multiple plant species & \multicolumn{2}{l}{\cite{dai2020agricultural}}                     \\ \cline{2-5} 
\multicolumn{1}{l|}{}   & DATF-GAN                & Multiple plant species & \multicolumn{2}{l}{\cite{dai2020crop}}                     \\ \cline{2-5} 
\multicolumn{1}{l|}{}   & WGAN-GP                 & Multiple plant species & \multicolumn{2}{l}{\cite{bi2020improving}}                        \\ \cline{2-5} 
\multicolumn{1}{l|}{}   & DoubleGAN               & Multiple plant species & \multicolumn{2}{l}{\cite{zhao2021plant}}                     \\ \cline{2-5} 
\multicolumn{1}{l|}{}   & RAHC\_GAN               & Tomato leaves          & \multicolumn{2}{l}{\cite{deng2021rahc_gan}}                    \\ \cline{2-5} 
\multicolumn{1}{l|}{}                         & Re-enforcement GAN                           & Multiple plant species & \multicolumn{2}{l}{\cite{nerkar2021cross}} \\ \cline{2-5} 
\multicolumn{1}{l|}{}                         & Style consistent image translation            & Tomato leaves          & \multicolumn{2}{l}{\cite{xu2021style}}         \\ \hline
\end{tabular}
}
\end{table*}

\subsubsection{Weed Control} \label{sec:5.1.2}
Weeds compete with crops for resources (i.e., water and nutrients) and provide hosts for pests and diseases, posing a vital threat to crop production \citep{chen2021performance}. Machine-vision-based weed control offers a promising means for managing weeds effectively, especially mitigating herbicide resistance, by identifying and localizing weed plants followed by site-specific, individualized treatments (e.g., spot spraying, killing weeds by high-flame laser). Currently it remains a critical challenge to develop machine vision systems capable of accurate weed recognition and robust to variable field light conditions \citep{chen2021performance,westwood2018weed}. 

\cite{fawakherji2021multi} applied CGAN (Section~\ref{sec:4.1}) to generate four-channel multispectral (RGB + NIR) images for crop/weed segmentation. Instead of pursuing a full-scene image generation, the authors conditioned GAN models on the shape/mask of plant regions to produce semi-artificial images by replacing only targeted plant objects in images with synthetic instances, leaving soil backgrounds unchanged. A spatially adaptive denormalization GAN architecture that synthesizes photo-realistic images given semantic layout input \citep{park2019semantic} was adapted to perform semantic image synthesis. Evaluation on three public datasets, i.e., two sugar beet datasets \citep{chebrolu2017agricultural} and the sunflower dataset created by the authors, showed that the GAN-augmented data could statistically improve pixel-wise segmentation accuracies compared to the original data, with improvements in mean IoU (mIoU) of 0.03-0.10 depending on datasets and segmentation networks. The ensemble of RGB and NIR images yielded consistently better performance than using RGB conventionally. \cite{espejo2021combining} used DCGAN (section~\ref{sec:4.2}) to generate synthetic tomato and black nightshade images for weed identification tasks on the early crop dataset (\url{https://github.com/AUAgroup/early-crop-weed}). The FID metric \citep{heusel2017gans} was employed to determine the best DCGAN configuration for image generation. Synthetic tomato and black nightshade images generated were added to the original images and transfer learning was adopted to train several CNN classifiers (e.g., Xception, DenseNet and Inception). The Xception network yielded an optimal F1-score of 99.07\% using GAN-augmented training dataset compared to 98.22\% of training with traditional augmentation methods.

Unmanned aerial vehicles (UAV) or drone-based imagery has been widely used in precision farming to detect plant health conditions including weed infestations. Remote sensing by UAVs coupled with DL algorithms enables field-scale weed recognition and monitoring to support site-specific management. To address data insufficiency in weed-crop classification tasks, \cite{kerdegari2019semi} applied a semi-supervised GAN (SS-GAN) model adapted from DCGAN for pixel-wise classification of multispectral images acquired by a UAV. Unlike typical GAN where the \textit{discriminator} is a binary classifier for discriminating real and fake images, the SS-GAN used a multi-class classifier \textit{discriminator} taking the input of synthetic and label samples as well as unlabeled. Evaluated on the weedNet \citep{sa2017weednet}, a multispectral (RGB+NIR) dataset acquired by a UAV from sugar beet fields, the GAN model achieved F1 scores of about 0.85 by using two-channel (Red+NIR) images with 50\% labeled data \citep{kerdegari2019semi}. \cite{khan2021novel} employed SS-GAN for classifying crops and weeds in the RGB images acquired by UAVs from pea and strawberry fields at early crop growth stages. The SS-GAN achieved an overall classification accuracy of about 90\% when only 20\% labeled samples were used for training, which compared favorably to the results obtained by conventional supervised classifiers (e.g., K-nearest neighbor, SVM and CNNs). 

\subsubsection{Fruit Detection} \label{sec:5.1.3}
In-orchard fruit detection is challenging due to unstructured environments and variable field light. Over the past few years, DL-based object detectors and segmentation networks have been widely researched for fruit detection towards robotic harvesting, fruit counting and yield estimation \citep{koirala2019deep,maheswari2021intelligent}. To improve fruit recognition performance, GANs (Table~\ref{tab4}) have been utilized to generate realistic images for training DL models while reducing the requirements of manual efforts for data collection and labeling.


\cite{barth2020optimising} used CycleGAN to perform domain adaptive semantic segmentation to support robotic perception. The Capsicum annuum dataset \citep{barth2018improved}, consisting of 50 images of pepper plants collected in a greenhouse and 10500 synthetic images rendered in Blender (open-source 3D modeling software), was used to train CycleGAN for domain translation (\textit{synthetic} domain to \textit{empirical} domain). The mean color distribution correlation with the empirical data was improved from 0.62 (on empirical data) to 0.90 (after translation). Using translated synthetic images for part segmentation gave an 8\% IoU improvement by fine-tuning with empirical images compared to only using synthetic images, and a 55\% improvement compared to training without empirical fine-tuning.  \cite{luo2020pine} reported on the detection of pine cones using boundary equilibrium GAN (BEGAN) \citep{berthelot2017began} and YOLOv3 \citep{redmon2018yolov3}. The BEGAN uses an equilibrium enforcing method to control the tradeoff between image diversity and visual quality. The authors used the GAN-generated pinecones instances to replace pine cones in the original images, which resulted in the detection average precision (AP) increasing to 95.3\% from 93.4\% for the model without image augmentation. 

\cite{bellocchio2020combining} used CycleGAN to perform image domain translation coupled with weakly-supervised learning for fruit counting where only fruit presence-absence labels are given. Experiments on four orchard datasets showed CycleGAN could adapt image conditions (e.g., illumination, blur, saturation) to the target domain and transform fruit colors and textures from a source orchard to a different target orchard. \cite{fei2021enlisting} used CycleGAN with semantic constraints for domain adaptation of grape images while preserving spatial semantics such as fruit position and size. The GAN model was trained to adapt source domain vineyard images that were 3D rendered by Helios \citep{bailey2019helios}, to day and night images (target domains) collected from the orchard. Fruit detection models with YOLOv3 yielded 37.2 mAP@IoU0.5 by fine-tuning the detection model with the GAN synthesized images for day domain, compared to 37.0 mAP@IoU0.5 without the image synthesis, and 26.7 AP@IoU0.5 compared to 22.8 AP@IoU0.5 for night domain. It is noted that, although realistic images were generated by the GAN model, further validations are needed on using physically collected real-world data as source domain images.

Fruit occlusions in canopy pose great challenges to accurate fruit detection/localization and crop yield estimation.  \cite{olatunji2020reconstruction} applied CGAN to reconstruct complete kiwifruit surfaces by translating occluded fruit into non-occluded images to address partial occlusion problems in fruit detection. By reconstructing missing surface information of occluded fruit, the fruit shape, surface area and weight could be predicted from the reconstructed images. The CGAN was trained on 3000 image pairs computationally created in Blender, for generating the complete fruit surface topography. Experimental results on the synthetic validation dataset show that 75\% of the volume prediction had an error rate less than 5\% and only 7\% of volume predictions had error rate larger than 10\%. The CGAN model was able to generate high-quality images and yielded a predicted fruit weight error of 5.58\% validated on a dataset collected by a 3D scanner posed at a single angle to mimic occlusion. The authors however did not discuss the impact on model performance with different numbers of generated training images. Similar to \cite{olatunji2020reconstruction},  \cite{kierdorf2021behind} applied Pix2Pix \citep{isola2017image} to translate occluded berry images due to leaves to non-occluded fruit images for estimating the number of berries, enabling better yield estimation.

\begin{table*}[!ht]
\renewcommand{\arraystretch}{1.4}
\centering
\caption{Applications of GANs in weed control, aquaculture, fruit recognition, and livestock farming.}
\label{tab4}
\resizebox{0.65\textwidth}{!}{
\begin{tabular}{l|l|l|ll}
\hline
Application(s) & GAN(s)              & Material(s)        & \multicolumn{2}{l}{References}                  \\ \hline
\multirow{4}{*}{Weed Control}    & CGAN/CDCGAN & Crop/weeds           & \multicolumn{2}{l}{\cite{fawakherji2021multi}} \\ \cline{2-5} 
               & DCGAN               & Crop/weeds         & \multicolumn{2}{l}{\cite{espejo2021combining}} \\ \cline{2-5} 
               & Semi-Supervised GAN & Crop/weeds         & \multicolumn{2}{l}{\cite{kerdegari2019semi}}     \\ \cline{2-5} 
               & DCGAN               & Crop/weeds         & \multicolumn{2}{l}{\cite{khan2021novel}}          \\ \hline
\multirow{7}{*}{Fruit Detection} & CycleGAN    & Sweet or bell pepper & \multicolumn{2}{l}{\cite{barth2020optimising}}      \\ \cline{2-5} 
               &   CycleGAN       & Sweet or bell pepper & \multicolumn{2}{l}{\cite{barth2018improved}}         \\ \cline{2-5} 
               & BEGAN               & Pine cones         & \multicolumn{2}{l}{\cite{luo2020pine}}           \\ \cline{2-5} 
               & CycleGAN           & Orchard            & \multicolumn{2}{l}{\cite{bellocchio2020combining}}    \\ \cline{2-5} 
               & CycleGAN            & Vineyard           & \multicolumn{2}{l}{\cite{fei2021enlisting}}           \\ \cline{2-5} 
               & CGAN                & Kiwi fruits        & \multicolumn{2}{l}{\cite{olatunji2020reconstruction}}     \\ \cline{2-5} 
               & CGAN/CDCGAN         & Grapevine berries  & \multicolumn{2}{l}{\cite{kierdorf2021behind}}      \\ \hline
\multirow{2}{*}{Aquaculture}     & DCGAN       & Fish species         & \multicolumn{2}{l}{\cite{zhao2018semi}}       \\ \cline{2-5} 
               & CycleGAN            & Shrimp eggs        & \multicolumn{2}{l}{\cite{zhang2021shrimp}}            \\ \hline
\multirow{3}{*}{Animal Farming}  & SAGAN       & Goats                & \multicolumn{2}{l}{\cite{li2020dairy}}         \\ \cline{2-5} 
               & CGAN                & Cattle Muzzle      & \multicolumn{2}{l}{\cite{singh2021muzzle}}         \\ \cline{2-5} 
               & CTGAN               & Poultry Chicken   & \multicolumn{2}{l}{\cite{ahmed2021approach}}         \\ \hline
\end{tabular}
}
\end{table*}

\subsubsection{Aquaculture} \label{sec:5.1.4}
Application of imaging technology in precision aquaculture farming faces a number of special challenges \citep{li2021automatic, fore2018precision} in detecting and monitoring aquatic species under adverse poor underwater conditions (e.g., poor illumination and object visibility in turbid water, cluttered background) that make acquiring high-fidelity/-contrast images difficult. The insufficiency of aquaculture images available would add to the complexity of underwater species recognition tasks.


GANs have been applied for image synthesis for enhancing visual recognition in aquaculture (Table~\ref{tab4}). \cite{zhao2018semi} used a semi-supervised learning model based on the modified DCGAN (with a multi-class classifier in the \textit{discriminator}) for live fish recognition, on two open-sourced fish datasets, i.e., the Fish4-knowledge \citep{boom2012supporting} project (27370 fish images in 23 clusters) and the Croatian fish dataset \citep{jager2015croatian} (794 images of 12 fish species). Using small fractions (5\%-15\%) of labeled training samples, the semi-supervised GAN model achieved fish identification accuracies of 80.52\%-83.07\%, exceeding the accuracies obtained by CNN-based models \citep{marburg2016deep,qin2016stacking} without data augmentation as well as those based on other GANs (e.g., 48.39\%-59.44\% for the standard DCGAN). Improvements were also observed with higher fractions (50\%-75\%) of the labeled Croatian fish data used for fish identification. \cite{zhang2021shrimp} used CycleGAN \citep{Zhu_2017_ICCV} to generate images for automatic shrimp egg counting which otherwise would be labor-intensive in the hatcheries. A shrimp egg dataset was collected and manually labeled, containing 450 color images with about 272000 egg annotations. Randomly generated binary density maps combined with real images without annotations were used to train CycleGAN to generate photo-realistic images with precise egg annotations. A shrimp egg counting network combined with the fully convolutional regression network was proposed to count eggs through regressing the input image into its density map. Pretrained on the synthetic images and fine-tuned on the collected data, the network achieved the average counting accuracy of up to 99.2\%, compared to the accuracy of 96.6\% by training the synthetic data alone and 98.9\% only on the collected dataset.

\begin{figure*}[!ht]
  \centering
  \includegraphics[width=0.5\textwidth]{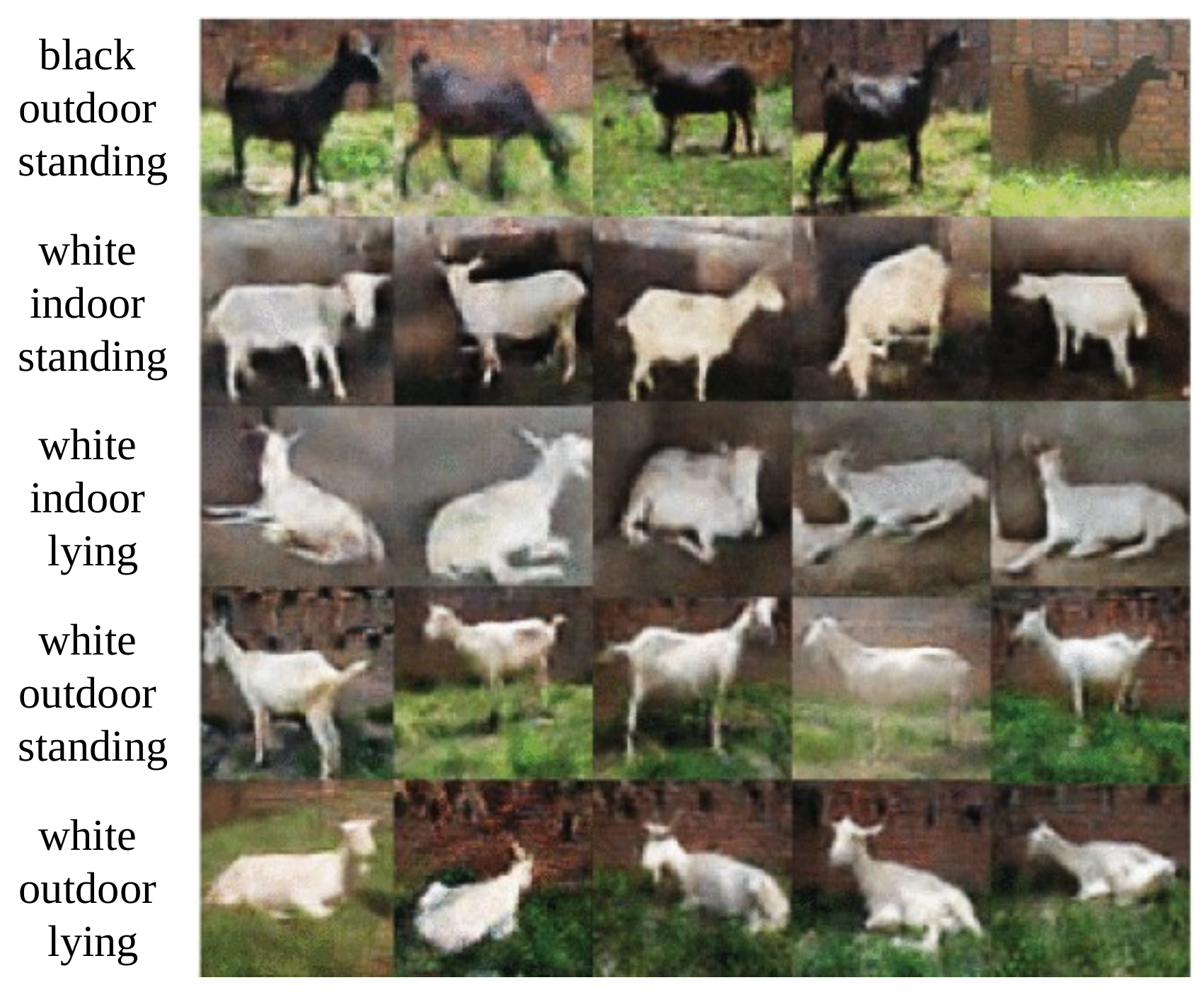}
  \caption{Dairy goat images generated by improved SA-GAN with corresponding multi-labels \citep{li2020dairy}.}
  \label{fig18}
\end{figure*}

\subsubsection{Animal Farming} \label{sec:5.1.5}
Animal farming here refers to livestock and poultry farming but excludes aquaculture described above. Computer vision systems assisted by DL algorithms have been investigated in animal farming for phenotyping and behavioral identification and monitoring \citep{wurtz2019recording, junior2021potential}, at group or individual levels, to enhance production efficiency a well as animal welfare, with applications expanded recently \citep{junior2021potential, li2021evaluating}. However, the exploitation of DL can be hurdled by the drudgery of collecting and labeling large amounts of animal image data. GANs are thus attractive because of their ability to generate images and labels automatically to empower DL models.      

\cite{li2020dairy} evaluated an improved SA-GAN (Section~\ref{sec:4.9}) that used a normalized self-attention mechanism and replaced one-hot labels with multi-class labels (colors, backgrounds, and behaviors) to generate high-quality dairy goat images (Fig.~\ref{fig18}). Experiments on a collection of goat images confirmed the effectiveness of the GAN model for enhancing image quality compared to image generation using one-hot labels. \cite{singh2021muzzle} applied CGAN to generate enhanced images for cattle identification based on muzzle pattern images. Low-quality images that were obtained by artificially degrading the original images, alongside binarized muzzle pattern images (the target enhanced images), were used to train the GAN model to learn image enhancement mapping. The synthetic images coupled with feature extraction and matching achieved an identification accuracy of 98.86\% for four breeds of cattle. This study, however, did not report cattle identification accuracy based on the original images. Wearable sensors have been widely researched for animal tracking and monitoring the behaviors of individuals \citep{siegford2016assessing}. In \cite{ahmed2021approach}, the time-series data collected by activity sensors placed on chicks \citep{abdoli2018time} was examined for classifying healthy and sick chickens. A GAN method designed for modeling tabular data, i.e., conditional tabular GAN \citep{xu2019modeling}, was used to expand the original sensor data for the classification, resulting in the accuracy of 97\% by TabNet \citep{siegford2016assessing}. Currently, the applications of GANs are still scant for synthesizing image data in poultry farming.

\subsection{Plant Phenotyping} \label{sec:5.2}
Plant phenotyping/phenomics in crop breeding is to quantify various plant phenotypes (e.g., growth dynamics, stress resistance) resulting from the interactions of genetics and environmental conditions. Imaging technologies play a crucial role in the realization of high-throughput, automated quantification of plant phenotypes \citep{minervini2015image,das2019leveraging}, thereby accelerating breeding processes and bridging the genotype-phenotype gap \citep{furbank2011phenomics}. Imaging-based plant phenotyping is still facing challenges with extracting meaningful phenotypic information from images due to factors such as lighting variations, plant rotations, occlusions \citep{das2019leveraging}. Data-driven machine learning methods have been leveraged by plant scientists for effective feature extraction, plant trait identification/classification and quantification \citep{singh2018deep,jiang2020convolutional}. To harness DL in plant phenotyping requires the curation of sufficient, high-quality labeled training samples, where GANs have emerged as powerful methods for synthetically expanding image datasets (Table~\ref{fig5}).


\cite{gulrajani2017improved} presented one of the earliest studies on image synthesis for plant phenotyping. The authors adapted DCGAN \citep{radford2015unsupervised} with the number of leaves of a plant as the conditional input, trained on the CVPPP2017 dataset of potted Arabidopsis plants (\url{https://www.plant-phenotyping.org/CVPPP2017}), which was able to generate realistic $128 \times 128$ color images. Later, \cite{zhu2018data} reported on using Pix2Pix  \citep{isola2017image} to synthesize plant images of $512 \times 512$ pixels for leaf counting. The GAN model was trained on a collection of 500 images from the CVPPP 2017 dataset, along with corresponding leaf segmentation masks. The authors evaluated their approach by applying Mask-RCNN \citep{he2017mask} on the augmented dataset, obtaining 16.67\% reduction in the leaf counting error compared to training the network on the original dataset without data augmentation. These two studies are however limited to generating samples for a single plant species. \cite{madsen2019generating} applied a Wasserstein auxiliary classifier GAN (WacGAN) that combined AC-GAN \citep{odena2017conditional} and WGAN-GP \citep{gulrajani2017improved}, to generate plant seedling images of multiple species. The WacGAN yielded 58.9±9.2\% plant recognition accuracy on the 9-species plant seedling dataset \citep{Giselsson2017}, evaluated using ResNet-101 \citep{he2016deep}. Subsequently, based on WacGAN \citep{madsen2019generating} and InfoGAN \citep{chen2016infogan}, \cite{madsen2019disentangling} presented WacGAN-info that incorporates unsupervised latent input variables in the GAN configuration for generating plant seedling images, achieving an improved classification accuracy of 64.3\%. However, the GAN model produced low-resolution images of $128 \times 128$ pixels, and its impact on plant classification was not examined in comparison with training on either the original data without data augmentation or images augmented by traditional methods. 

To deal with insufficient data in plant vigor rating, \cite{zhu2020data} applied conditional DCGAN (cDCGAN) to the images of orchid seedlings of two vigor ratings (healthy and weak). The \textit{generator} and \textit{discriminator} networks were conditioned by concatenating one-hot class labels at all layers for image generation, and the classification using ResNet-50 \citep{he2016deep} with the augmented data achieved a 23\% increase in the testing F1 score compared to the model trained without data augmentation. The authors also showed that adding bypass deconvolutional connections in the GAN architecture helped generate more realistic images. \cite{drees2021temporal} assessed the utility of CGAN based on Pix2Pix \citep{isola2017image} for predicting plant images at future growth stages. The GAN model was evaluated on two open-source datasets, i.e., the Arabidopsis plant dataset collected indoors \citep{bell_jonathan_2016_168158} and the field-collected cauliflower dataset \citep{bender2020high}. For instance segmentation of plants by Mask-RCNN \citep{he2017mask}, the correction between the size of predicted and reference images reached $R^2 = 0.95$ for the cauliflower dataset but decreased to the values of 0.66-0.82 for the second dataset. Several factors influenced the image generation, such as geo-reference of images (required for Pix2Pix image alignment), plant position and overlaps of neighboring plants, which restrict the applicability of Pix2Pix-based GANs to complex field images.

Collecting large-scale datasets from diverse fields for plant phenotyping is a lengthy process and resource intensive.  Existing phenotyping datasets mostly consist of only images acquired under well-controlled laboratory conditions. To address the paucity of field-based data, \cite{shete2020tasselgan} proposed TasselGAN modified from DCGAN \citep{radford2015unsupervised} to generate images of maize tassels against synthetic sky backgrounds. The authors generated maize tassel images from the indoors collected data \citep{gage2017tips} and synthetic sky background patches using the data in \citep{dev2016color,li2011hybrid} separately, and then merged them to form field-like images. However, the perceptual quality of generated images needs improvements, given the fact less than 40\% of the generated images were considered sufficiently realistic based on qualitative assessment. The resultant image resolution ($128 \times 128$ pixels) is not well suitable for phenotyping complex plant traits.  An open-source field dataset called Global Wheat Head Dataset (GWHD) \citep{david2020global}, comprising 4700 high-resolution RGB images and 190000 labeled wheat heads collected from 12 countries, was recently released to facilitate wheat phenotyping research, particularly on the detection of wheat heads. The dataset, however, has domain or distribution shift issues \citep{koh2021wilds,zhu2018data} with substantial variations across growth region and crop varieties, which can degrade the performance of machine learning models. While simulated or synthetic data can alleviate the issue \citep{najafian2021semi}, a domain gap exists between the simulated and read images. To tackle domain shifts, \cite{hartley2021domain} used CycleGAN \citep{Zhu_2017_ICCV} to perform \textit{synthetic to real} domain adaptation for the GWHD dataset \citep{david2020global}. The GAN model was trained using a synthetic wheat head dataset, containing over 5000 images with over 100000 annotations, and part of the GWHD data as source and target domains, respectively. The wheat head detection was done by Mask-RCNN via Detectron2 \citep{wu2019detectron2} and mIoU gains of about 3-4\% were achieved by using the combination of synthetic and real samples compared to baseline models using real wheat head data alone. Only using synthetic data, however, lead to dramatically deteriorated accuracy, highlighting the value of labeled real images.

\begin{table*}[!ht]
\centering
\renewcommand{\arraystretch}{1.4}
\caption{Application of GANs in plant phenotyping and post-harvest quality assessment.}
\label{tab5}
\resizebox{0.75\textwidth}{!}{
\begin{tabular}{l|l|l|ll}
\hline
Application                             & GAN   Variant     & Materials   Examined                           & \multicolumn{2}{l}{Reference(s)}             \\ \hline
\multirow{10}{*}{Plant Phenotyping}     & DCGAN             & Arabidopsis plants                             & \multicolumn{2}{l}{\cite{valerio2017arigan}} \\ \cline{2-5} 
                                        & CGAN (Pix2Pix)    & Arabidopsis plants                             & \multicolumn{2}{l}{\cite{zhu2018data}}        \\ \cline{2-5} 
                                        & WacGAN            & Multiple species plant seedlings               & \multicolumn{2}{l}{\cite{madsen2019generating}}   \\ \cline{2-5} 
                                        & WacGAN-Info       & Multiple species plant seedlings              & \multicolumn{2}{l}{ \cite{madsen2019disentangling}}    \\ \cline{2-5} 
                                        & cDCGAN            & Orchid vigor rating                            & \multicolumn{2}{l}{\cite{zhu2020data}}       \\ \cline{2-5} 
                                        & CGAN (Pix2Pix)    & Arabidopsis plants and field-grown cauliflower & \multicolumn{2}{l}{\cite{drees2021temporal}}      \\ \cline{2-5} 
                                        & TasselGAN (DCGAN) & Maize tassels                                  & \multicolumn{2}{l}{\cite{shete2020tasselgan}}      \\ \cline{2-5} 
                                        & CycleGAN          & Wheat heads                                    & \multicolumn{2}{l}{\cite{hartley2021domain}}       \\ \cline{2-5} 
                                        & TasselGAN (DCGAN) & Maize tassels                                  & \multicolumn{2}{l}{ \cite{shete2020tasselgan}}      \\ \cline{2-5} 
                                        & CycleGAN          & Wheat heads                                    & \multicolumn{2}{l}{\cite{hartley2021domain}}       \\ \hline
\multirow{6}{*}{Postharvest assessment} & Customized GAN    & Coffee beans                                    & \multicolumn{2}{l}{\cite{chou2019deep}}       \\ \cline{2-5} 
                                        & DCGAN             & Jujubes                                         & \multicolumn{2}{l}{\cite{guo2021quality}}        \\ \cline{2-5} 
                                        & CGAN              & Lemons                                          & \multicolumn{2}{l}{\cite{DBLP:journals/corr/abs-2104-05647}}       \\ \cline{2-5} 
                                        & CoGAN             & Potatoes                                        & \multicolumn{2}{l}{\cite{marino2020unsupervised}}     \\ \cline{2-5} 
                                        & TransGAN             & Lychees                                  & \multicolumn{2}{l}{\cite{wang2021lychee}}      \\ \cline{2-5} 
                                        & VanillaGAN          & Wheat kernels                                         & \multicolumn{2}{l}{\cite{yang2021detection}}     \\ \hline
\end{tabular}
}
\end{table*}

\subsection{Postharvest Quality Assessment} \label{sec:5.3}
Machine vision or imaging technologies are pervasive for postharvest quality assessment of agricultural products \citep{CHEN2002173,Blasco2017, LU2020111318}. At modern packing facilities, machine vision (color or monochromatic imaging) systems have been fairly well established for automated grading and sorting  of food products for quality such as shape, size, and color. Because of significant biological variations, tissue defects are however still manually inspected for many horticultural products. DL algorithms hold promise to empower machine vision systems with enhanced capabilities for defect detection of agricultural produce. Several studies (Table~\ref{tab5}) have been performed recently on the use of GANs for synthesizing fruit images for defect detection described below. 

\cite{chou2019deep} used GAN in a DL pipeline for identifying defective coffee beans. The defect inspection module was comprised of YOLOv3 \citep{redmon2018yolov3} and Hough circle detectors combined with a GAN-structured image generation mechanism, which could achieve detection rates of up to 80\% for defective beans. However, the pixel resolution of generated images is still limited in this study, which can be detrimental to tasks like segmentation and quantification of defect areas. \cite{guo2021quality} used DCGAN to generate jujube images for quality grading (1 healthy and 6 defective classes). Images ($256 \times 256$ pixels) for defect classes were synthesized to alleviate the imbalance between healthy and defective samples. The inclusion of synthetic samples by DCGAN achieved classification accuracies of up to 98\%, representing performance gains of 2-4\% compared to models based on traditional data augmentation. It is noted that the GAN images for each defect class were generated separately rather than through a multi-class data generation process, which can be potentially more efficient. Recently, \cite{DBLP:journals/corr/abs-2104-05647} applied CGAN to generate synthetic lemon images for classifying healthy and unhealthy fruit, using a public lemon dataset \citep{softwaremill_2020}. By using synthetic fruit images, the classification based on VGG16 achieved an accuracy of 88.75\% against 83.77\% without data augmentation. With a varied number (200 – 3,000) of synthetic images for model training, the best accuracy was attained by using only 400 images ($\sim$13.5\% of the whole dataset). The authors did not examine reasons for this phenomenon, which may be due to the quality of generated images or the capacity of classification models.

In real-world applications, imaging systems may be subjected to modifications (e.g., lighting conditions) over time, which can lead to dataset shift issues \citep{quinonero2008dataset}. To address dataset shift problems in potato quality control, \cite{marino2020unsupervised} reported an unsupervised adversarial domain adaptation  (ADA) technique, which integrates adversarial learning and domain adaptation similarly to GNAs, for potato defect classification. Two datasets were collected for white potatoes and red tubers of six classes (1 healthy and 5 defective), respectively. The authors conducted two domain adaptation  tasks, i.e., simulating lighting condition changes on the target domain by artificially increasing the brightness of part of white potato images and translating white potato to red tuber images. In the framework of adversarial discriminative domain adaptation  \citep{tzeng2017adversarial}, their ADA method with GoogleNet as the target classifier achieved significant performance gains on the target domain, with an overage F1-score increasing to 0.84 from 0.46 in case of no domain adaptation . 

Transformer-based GANs have been proposed recently for improving the state of the art in image generation \citep{jiang2021transgan, hudson2021generative, zhao2021improved}. \cite{wang2021lychee} reported on using TransGAN \citep{jiang2021transgan} (Section~\ref{sec:4.11}) to generate lychee images for surface defect detection. Images were preferentially generated for two defect classes to rebalance training data, and the augmented-data achieved improvements of 0.58\%-2.86\% in mAP for defect detection, depending on object detectors, compared to training without GAN-based data augmentation. It is noted that the authors included only part of generated images for model training, because TransGAN did not always generate high-quality images. GANs have also been employed for generating non-imaging sensor data for quality assessment of agricultural products. In \cite{yang2021detection}, 2D spectrograms of impact acoustic signals collected with a microphone were used for classifying wheat kernels (two defect classes along with healthy kernels). The authors used a vanilla GAN to generate synthetic spectrograms for data augmentation, obtaining an F1 score of 96.2\% by using a PANSNet-5 network \citep{liu2018progressive}. This study however did not examine performance gains due to the GAN-generated data, which would require ablation studies without using GAN samples.

\section{Challenges of GANs}\label{sec:dis}
Although GANs have demonstrated impressive performance in generating realistic images and improving DL models in agricultural applications, there still remain several open challenges with training and evaluation, as identified and discussed below, alongside potential opportunities, which need to be well understood for exploiting the benefits of GANs in agriculture.  

\subsection{Training Instability} \label{sec:6.1}
Training GANs is prone significantly to instability problems \citep{arjovsky2017towards, DBLP:journals/cacm/GoodfellowPMXWO20}, despite a set of regularization techniques proposed for improving training stability (e.g., model architectures, loss function modifications, gradient penalty, spectral normalization) \citep{wiatrak2019stabilizing, liu2021generative}. Two common failures of training include vanishing gradients \citep{goodfellow2016nips}, leading to convergence failures, and mode collapse or dropping \citep{goodfellow2016nips, durall2020combating, srivastava2017veegan}, where the \textit{generator} fails to learn well the distribution of training datasets.

GAN solves the minmax game (Eq.~\ref{eqn1}) by achieving the Nash equilibrium through a gradient descent method. When the \textit{generator} is not as good as the \textit{discriminator}, the \textit{discriminator} always differentiates between generated and real samples and the gradients of the \textit{generator} will vanish (close to 0), slowing or completely stopping the learning. On the other hand, mode collapse often happens, appearing as that the training data cannot be generated by the \textit{generator}. This is probably the biggest issue with GANs \citep{goodfellow2016nips}. The \textit{generator} may produce only few modes of training samples (partial mode collapse), and in the worst cases, simply the same set of instances (complete mode collapse) \citep{adler2018banach,arjovsky2017wasserstein,arjovsky2017towards}. Mode collapse restricts the ability of GAN to generate diverse images and is very detrimental for visual recognition tasks.

Currently even the state-of-the-art GANs (e.g., BigGAN and TransGAN) would suffer from instability issues when trained on complex datasets and cannot be guaranteed to consistently produce images with high enough quality \citep{wang2021lychee,yuwana2020data,espejo2021combining,shete2020tasselgan}. Moreover, the performance of GANs is generally sensitive to the settings of hyper-parameters \citep{kurach2019large,espejo2021combining}. Hence to make GANs work often requires significant manual efforts on hyperparameter tuning, network structure engineering, and a number of non-trivial training tricks \citep{kurach2019large, lucic2018gans}. This is conceivably true for agricultural applications where GANs are used for modeling the distribution of highly complex, multi-class/-modal datasets, especially for those collected from unstructured, cluttered field conditions. Additionally, imbalanced training data, which are frequently encountered in agricultural applications, can contribute to GAN training difficulty and mode collapse \citep{dieng2019prescribed, douzas2018effective, mariani2018bagan}.   


\subsection{GAN Evaluation} \label{sec:6.2}
Performance evaluation of GANs is another daunting task \citep{goodfellow2016nips}. Currently there are no standardized, universal metrics for evaluating GANs and the quality of generated samples. Conventional error metrics (e.g., root mean square error) are not suitable because of a lack of direct one-to-one correspondence between generated and real images. Qualitative, visual assessment of the perceptual quality of generated images is commonly performed in computer vision literature \citep{ledig2017photo, zhou2019hype}, but it is subjective, prone to human error, time-consuming for large datasets, and may not capture distributional characteristics of datasets. \cite{borji2019pros,borji2022pros} reviewed quantitative metrics for GAN evaluation, among which Inception Score (IS) \citep{salimans2016improved}, F1-score, Precision and Recall \citep{lucic2018gans}, Fréchet Inception Distance (FID) \citep{heusel2017gans}, mode score \citep{che2016mode} and the metrics for image structural similarity assessment \citep{wang2004image} are presumably the most popular for assessing the similarity between real and generated images. All of them, however, have shortcomings \citep{xu2018empirical,borji2019pros,borji2022pros}. There are also other metrics proposed for assessing the fidelity and diversity characteristics of generated images \citep{naeem2020reliable, kynkaanniemi2019improved}. For agricultural applications reviewed in Section~\ref{sec:app_gans}, only a handful of studies \citep{nazki2020unsupervised,shete2020tasselgan,espejo2021combining} reported on quantitative metrics for assessing GAN-generated images. The validity of these metrics for agricultural images remains to be explored at large. 

Instead of quantitatively evaluating the quality of generated images, most applications in Section~\ref{sec:app_gans} examined the impact on of GANs on the eventual performance of DL models in visual recognition tasks, in comparison to baseline models using traditional data augmentation or the original, un-augmented data. These studies provide overwhelming evidence that DL models substantially benefit from GAN-augmented data. For instance, \cite{xu2021style} observed the plant disease recognition rate increasing from 86.75\% of the original data to 94.61\% based on the ensemble of the original, GAN-synthesized images and the data from traditional augmentation. \cite{abbas2021tomato} obtained 97.11\% accuracy on GAN-augmented images compared to 94.34\% on original. \cite{sun2020data} reported performance gains of about 3\% due to GAN for plant disease classification. A more substantial performance gain of about 10\% was achieved in \cite{zhu2020data} on plant vigor rating. 

Although it seems to be easier to evaluate GANs on downstream machine learning tasks, quantitative evaluation of the quality and diversity of images generated by GANs is important for gaining an in-depth understanding of their strengths and limitations and inspire GAN innovations and should be explored in future research.

\subsection{Training with Limited Data} \label{sec:6.3}
High-quality image generation by GANs \citep{brock2018large, DBLP:journals/corr/abs-1907-02544,karras2019style,karras2020analyzing} is fueled by the seemingly sufficient supply of training images. For agricultural applications (e.g., plant disease detection) where there are only a small number of training images available, it can be difficult to train GANs to generate meaningful images beneficial for downstream DL model development. For instance, in \cite{zhu2020data} where GAN models were trained with varied numbers of training images to generate orchid seedlings, the generated images were perceptually inferior with the smallest number of training images, failing to capture detailed structure on leaves and roots. Compared to real samples, low-quality generated images may lack texture details and contain unrealistic, undesirable artifacts. In \cite{DBLP:journals/corr/abs-2104-05647} for generating lemon images, many synthetic images were found more reminiscent of potatoes than lemons and some suffered from unrealistic check-board patterns. Provision of sufficient samples for training GANs would facilitate generating high-quality, realistic images and eventually benefit DL models. 

The key issue with limited data for training GANs is fundamentally that the \textit{discriminator} overfits to training samples and consequently its feedback to the \textit{generator} become useless, leading to the divergence of the training process \citep{salimans2016improved,arjovsky2017wasserstein}. The traditional image augmentation methods described in Section~\ref{sec:basic_image_aug}, can be potentially used to prevent the \textit{discriminator} from overfitting and facilitate GAN training. In \cite{bi2020improving} on plant disease classification, traditional methods (e.g., flipping, rotation and contrast manipulation) were first used to augment the original data, followed by GAN training to generate synthetic images to train CNN models. This strategy was proven effective for achieving high image classification accuracies, which were also used in other studies \citep{arsenovic2019solving, douarre2019novel}. 

In addition, to train GANs with limited data, \cite{karras2020training} proposed an adaptive \textit{discriminator} augmentation mechanism that reliably stabilizes training and improves image generation of GANs when training data is in short supply. The authors also acknowledged that the augmentation should not be substituted for real data to train high-quality GANs. \cite{tran2021data} recognized the issues with classic data augmentation for GANs and proposed a principled framework of data augmentation optimized to improve the learning of \textit{discriminator} and \textit{generator}. These methods can be potentially useful for helping train GANs with limited data in agricultural applications.

\subsection{Other Considerations} \label{sec:6.4}
As discussed above, GAN models have the risk of generating low-quality images, while this may not be a real problem for traditional data augmentation methods that generally do not significantly impair image quality. That said, GAN-based data augmentation may not always translate into performance gains of DL models compared to traditional data augmentation or using the original data, which is observed in many application studies \citep{xu2021style,hartley2021domain,douarre2019novel,fawakherji2021multi}. However, strong evidence shows consistent improvements can be obtained by using both GAN and traditional augmentation than performing them separately, or by mixing the original and GAN-augmented data. This suggests that GAN and traditional image augmenation can work synergistically to improve the performance of DL models. More extensive research is still needed on systematic evaluation of the impact of GANs and traditional image augmentation in agricultural applications.   

Semi-supervised GANs aim to learn from data in a semi-supervised fashion where a small number of labeled images with large unlabeled data is to achieve model performance comparable to supervised training. \cite{kerdegari2019semi} successfully applied semi-supervised GAN to weed images with only half of labeled training data. Semi-supervised GANs should be well considered in case of the difficulty of labelling large-scale data. Makeup removal (removing irrelevant backgrounds or occlusions) is found effective for assisting in image generation of GANs. \cite{kierdorf2021behind} applied Pix2Pix GAN to remove the leaves from the berry images to generate non-occluded berry images for accurate berry counting. \cite{olatunji2020reconstruction} reconstructs the complete kiwifruit surface by translating occluded fruit into non-occluded images to address partial occlusion problems in fruit detection. More research on image processing like makeup removal can be beneficial to the application of GANs to visual recognition tasks like fruit detection and counting.

Multi-task learning is a modeling paradigm that enables simultaneously learning multiple related tasks with the goal to improve the generalization power for all the tasks \citep{zhang2021survey, vandenhende2021multi}. Currently, most GAN applications in agriculture focus on generating images for a single task (e.g., image classification  or object detection). Training in a multi-task manner can potentially foster generalization of GANs and improve overall performance for each task by leveraging knowledge contained in other tasks \citep{zhang2021survey, vandenhende2021multi}. A multi-task GAN model can be designed by using a multi-task \textit{discriminator} network. For instance, in \cite{bai2018sod}, an end-to-end multi-task GAN was proposed for detecting small objects, in which the \textit{discriminator} performs multiple tasks including distinguishing real from generated images, predicting object categories, and regressing bounding boxes simultaneously. This model showed superior object detection performance on the COCO dataset \citep{lin2014microsoft}. Currently there is scant research on multi-task GANs in agriculture, which would be worthy of investigation in agriculture application.

\section{Summary}
\label{sec:conclu}
Since its first proposal in 2014, GAN has attracted growing interest in computer vision areas. The ability of GANs to synthesize plausible, diverse images opens new opportunities to enable better-performing DL models trained for agricultural applications, especially when large-scale labeled image datasets are not readily available. This paper makes a first comprehensive review of GANs in agricultural contexts. We have given an overview of traditional image augmentation methods and the landscape of GAN architectural variants. Through a systematic review, we summarize a total of 59 publications of GANs in agriculture since 2017, for image augmentation in precision agriculture, plant phenotyping and postharvest inspection of agricultural products. GANs have achieved remarkable performance in various visual recognition tasks, such as plant disease classification, weed detection, fruit counting, postharvest fruit defect detection, etc. We also discuss challenges and opportunities regarding GAN training and evaluation. This paper will be beneficial for facilitating research and development of GANs in agriculture and food domains.     

\section*{Authorship Contribution}
\textbf{Ebenezer Olaniyi}: Formal analysis, Software, Writing - original draft;
\textbf{Dong Chen}: Formal analysis, Software, Writing - original draft; \textbf{Yuzhen Lu}: Conceptualization, Investigation, Supervision, Writing - original draft \& review \& editing; 
\textbf{Yanbo Huang}: Writing - review \& editing. 

\section*{Acknowledgement}
This work was supported in part by Cotton Incorporated award \#21-005 and the USDA National Institute of Food and Agriculture Hatch project \#1025922.

\typeout{}
\bibliography{ref}
\end{document}